\RequirePackage{fix-cm}
\documentclass{article} %
\usepackage[T1]{fontenc}
\usepackage{iclr2027_conference,times}
\usepackage{amsmath,amsfonts,bm}

\def\eqref#1{equation~\ref{#1}}

\def\1{\bm{1}}

\def\eps{{\epsilon}}

\DeclareMathAlphabet{\mathsfit}{\encodingdefault}{\sfdefault}{m}{sl}
\SetMathAlphabet{\mathsfit}{bold}{\encodingdefault}{\sfdefault}{bx}{n}

\newcommand{\E}{\mathbb{E}}

\usepackage{hyperref}
\usepackage{xurl} %
\usepackage{booktabs} 
\usepackage{graphicx}
\usepackage{xcolor}
\usepackage{caption}
\usepackage{wrapfig}
\usepackage{needspace}
\usepackage{enumitem}
\usepackage{tabularx,array}
\usepackage{longtable}
\usepackage{makecell}
\usepackage{multirow}
\usepackage{siunitx}
\usepackage{placeins}

\newlength{\figuretwotextsep}
\newlength{\figuretextsep}
\title{Residual Modeling Closes the Regression and Generative Policy Gap in Robot Learning\vspace{-0.2em}}

\author{
\makebox[\textwidth][c]{%
  \begin{tabular}{@{}cccc@{}}
    \textbf{Yuchen Zhou}$^{1,*}$ & \qquad
    \textbf{Jiacheng You}$^{2,*}$ & \qquad
    \textbf{Weikang Wan}$^{3}$ & \qquad
    \textbf{Weijun Dong}$^{3}$
  \end{tabular}
}
\\[0.5em]
\makebox[\textwidth][c]{%
  \begin{tabular}{@{}cc@{}}
    \textbf{Yang Gao}$^{2}$ &
    \hspace{2cm}\textbf{Jiayuan Mao}$^{1}$
  \end{tabular}
}
\\[0.7em]
\makebox[\textwidth][c]{%
  $^{1}$ University of Pennsylvania \qquad
  $^{2}$ Tsinghua University \qquad
  $^{3}$ UC San Diego
}
\\[-1.2em]
}

\iclrpreprintcopy %
\begin{document}

\vspace*{\dimexpr-3em-\baselineskip-1.67pt\relax}
\maketitle
\begingroup
\renewcommand{\thefootnote}{*}
\footnotetext{Equal contribution. Project Page: {\scriptsize \url{https://projects.thelabone.org/regression-policy/}}}
\endgroup
\suppressfloats[t]

\begin{abstract}
\vspace{-0.3em}
Learning from demonstration has enabled impressive robot behaviors.
A common choice for policy learning is to use diffusion or flow matching (Flow-Policies), which often outperforms direct action regression trained with mean squared error (MSE-Policies). This gap is commonly attributed to multimodal demonstrations.
We revisit this gap from the perspective of statistical modeling: how action-prediction residuals shape policy optimization.
Our analysis of real-world robot demonstration data reveals substantial state-dependent variation in residual scales and heavier-than-Gaussian tails.
While both MSE-Policies and Flow-Policies exhibit heavy-tailed action residuals, their training gradients behave differently: MSE allocates more gradient magnitude to observations with large action residuals, which hurts optimization. Motivated by these findings, we introduce heteroscedastic Student-$t$ action regression (HT-Policies), which learns input-dependent residual scales and reduces the influence of heavy tails.
HT-Policies predict action chunks with a single feed-forward pass and can reuse pretrained flow-matching-based policy networks as the backbone.
Across four simulation benchmarks and real-robot evaluations, HT-Policies achieves success rates competitive with generative policy baselines, both when trained from scratch and from pretrained vision-language-action and world-action models, despite being faster in training and inference.
Together, these findings shed light on the practical advantages of generative objectives in robot learning from demonstrations and offer an efficient direct-regression alternative for a range of architectures and tasks.
\end{abstract}
\vspace{-0.6em}

\vspace{-0.5em}
\section{Introduction}
\label{sec:introduction}
\vspace{-0.5em}

Learning robot policies from demonstrations is typically formulated as a supervised learning problem, in which an action chunk --- a short sequence of consecutive actions --- is predicted from observations and goal specifications~\citep{robomimic2021}. A common approach for policy learning is to use generative models, such as diffusion and flow-matching models~\citep{chi2023diffusionpolicy,pan2026much}. In practice, these models often outperform direct action regression trained with mean squared error (MSE-Policies). This performance gap is commonly attributed to multimodality: generative policies can model multiple action modes, whereas MSE-Policies tend to average across modes, potentially producing actions that do not correspond to any demonstrated behavior.

\begin{figure*}[t]
    \centering\small
    \vspace{-0.5em}
    \makebox[\textwidth][c]{%
        \includegraphics[width=\textwidth]{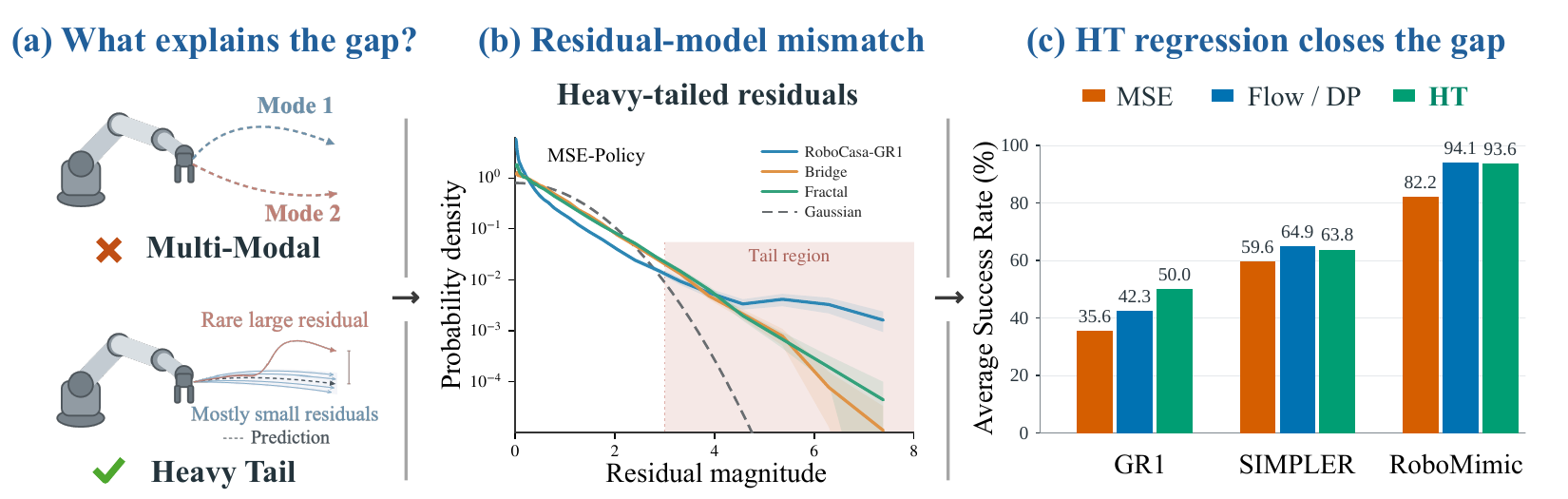}%
    }
    \vspace{-1em}
    \caption{
    \textbf{(a)} We attribute the performance gap between MSE- and Flow-Policies to different statistical modeling of prediction residuals.
    \textbf{(b)} We found observation-dependent and heavy-tailed residuals for MSE-Policies on multiple benchmarks.
    \textbf{(c)} Benchmark success rates (\%) for different policy classes.
    We use flow-matching for RoboCasa-GR1 and SIMPLER, and diffusion for RoboMimic.
    }
    \vspace{-0.5em}
    \label{fig:teaser}
\end{figure*}

Recent results suggest that explicitly modeling multimodal action distributions is not necessary for competitive performance. Minimal Iterative Policy (MIP), introduced by \citet{pan2026much}, uses a {\it deterministic} network in two successive passes for action prediction. The first pass predicts an action chunk conditioned only on the policy inputs; the second pass additionally conditions on the first prediction to refine it. During training, the network is supervised with a denoising objective. MIP closes the performance gap with Flow-Policies and attributes its success to iterative refinement, which helps keep predicted actions on the {\it manifold} of demonstrated actions. However, this explanation remains largely empirical, with limited formal characterization and real-data verification.

In this paper, we revisit this problem from a statistical modeling perspective. Specifically, we examine the residual models implicit in different training objectives and how they affect policy optimization. This perspective offers a way to understand the benefits of denoising supervision and motivates a direct action regression objective that can resemble those benefits without iterative inference.

We begin by examining the action residuals of trained policies. From a likelihood perspective, MSE regression corresponds to a fixed-scale Gaussian residual model. Empirically, however, both MSE-Policies and Flow-Policies exhibit heavy-tailed action residuals. One source of these tails is variation in residual scale across observations: for example, reaching toward a hole and making fine adjustments during insertion may involve different prediction-error scales. Yet scale variation alone does not fully account for the observed tails, which remain heavier than Gaussian after per-observation scale normalization, as illustrated in Figure~\ref{fig:teaser}b.
This structure matters for optimization because the gradient of MSE with respect to the predicted action is proportional to the residual. We show that high-residual samples account for a dominating share of gradient magnitude under MSE. Flow matching exhibits a similar concentration at high noise levels, but distributes gradient magnitude more evenly, especially at low noise levels due to its denoising training objective.

Motivated by these findings, we introduce \emph{heteroscedastic Student-$t$} action regression (HT-Policies)~\citep{lange1989robust}. HT-Policies incorporate the observed residual structure directly into the regression objective, predict action chunks with a single feed-forward pass, and can reuse pretrained flow-matching policy networks as backbones.
In four simulation benchmarks and real-robot evaluations, HT-Policies achieve success rates competitive with generative policy baselines. They perform well both when trained from scratch and when fine-tuned from pretrained vision-language-action and world-action models, while improving training and inference efficiency. Our gradient analysis shows reduced concentration of gradient magnitude on high-residual observations, supporting our hypothesis about the role of residual modeling in policy optimization. Together, these findings shed light on the practical advantages of generative objectives in robot learning and offer practitioners an efficient direct-regression alternative across a range of architectures and tasks. To summarize, our contributions are as follows:
\begin{itemize}[leftmargin=1.2em, itemsep=2pt, parsep=0pt, topsep=2pt]
  \item We characterize robot-policy residuals, revealing input-dependent scales and heavy tails that expose a mismatch with the fixed-scale Gaussian assumption of MSE.
  \item We further show that this mismatch causes MSE to overemphasize large-action-residual samples, whereas flow matching distributes gradients more evenly, especially in low-noise regimes. We explain the mechanism of this by studying the denoising objectives.
  \item We introduce HT-Policies, which explicitly model input-dependent scales and heavy tails. Across simulation and real-robot tasks, they match generative baseline performances while enabling more efficient training and inference.
\end{itemize}

\vspace{-1em}
\section{Related Work}
\label{sec:related-work}
\vspace{-0.5em}

\textbf{Variations in robot demonstration data.}
Human demonstrations exhibit variation arising from multiple sources, including differences in task strategies~\citep{shafiullah2022behaviortransformerscloningk,pmlr-v205-gandhi23a,jia2024towards}, action rates~\citep{shi2026diversity}, and execution errors or inaccuracies~\citep{Chen2025RestoringND}.
Prior work has addressed such variation from two main perspectives.
One line of work treats it as distributional multimodality and models it with Gaussian mixtures~\citep{robomimic2021}, diffusion models~\citep{chi2023diffusionpolicy}, or conditional VAEs~\citep{zhao2023learningfinegrainedbimanualmanipulation}.
Others treat variation as undesirable noise and seek to suppress it~\citep{shi2026diversity,Chen2025RestoringND}.
While these approaches can accommodate diverse demonstrations, they largely study how variation should be modeled or filtered empirically, rather than understanding its statistical structure and how it affects learning.

MIP~\citep{pan2026much} shows that averaging flow-policy outputs leaves performance largely unchanged, a finding we extend to VLA and WAM policies (Appendix~\ref{sec:appendix-flow-action-averaging}). Instead of avoiding mode averaging, it argues that flow's iterative refinement moves predictions toward the data manifold, but provides limited formal characterization. Moreover, its study of off-manifold error on out-of-distribution states does not distinguish manifold adherence from generalization. We revisit the gap by studying the statistical residual modeling in MSE-Policies and its effects on policy learning.

\textbf{Robust and heteroscedastic regression.}
Huber losses~\citep{Huber_1964} and Student-$t$
likelihoods~\citep{lange1989robust} improve regression robustness to large residuals; adaptive robust losses can learn the appropriate degree of robustness from data~\citep{barron2019generaladaptiverobustloss}.
Complementarily, heteroscedastic regression models input-dependent residual scales~\citep{NixPitfall}, although variance-based reweighting under Gaussian likelihoods can adversely affect mean estimation~\citep{seitzer2022on}.
These objectives have been widely explored in computer vision, including bounding-box regression~\citep{girshick2015fastrcnn} and depth prediction~\citep{kendall2017uncertaintiesneedbayesiandeep}.
In imitation learning, OpenVLA-OFT shows strong LIBERO performance with
$L_1$ regression~\citep{kim2025fine}, although $L_1$ still underperforms MSE on
three of four LIBERO suites when fine-tuning $\pi_0$~\citep{pan2026much}.
Robust classification objectives have also been studied~\citep{tangkaratt2021robustimitationlearningnoisy}.
We combine robust regression with input-dependent scale learning for action learning.
\vspace{-1em}
\section{Residual Structure and Optimization}
\vspace{-0.5em}
\label{sec:data-analysis}
\label{sec:heavy-tail}
\label{sec:gradient-imbalance}
\label{sec:tail-tolerance}
\label{sec:tail-origin}

\begin{figure}[tp]
    \centering\small
    \includegraphics[width=\linewidth]{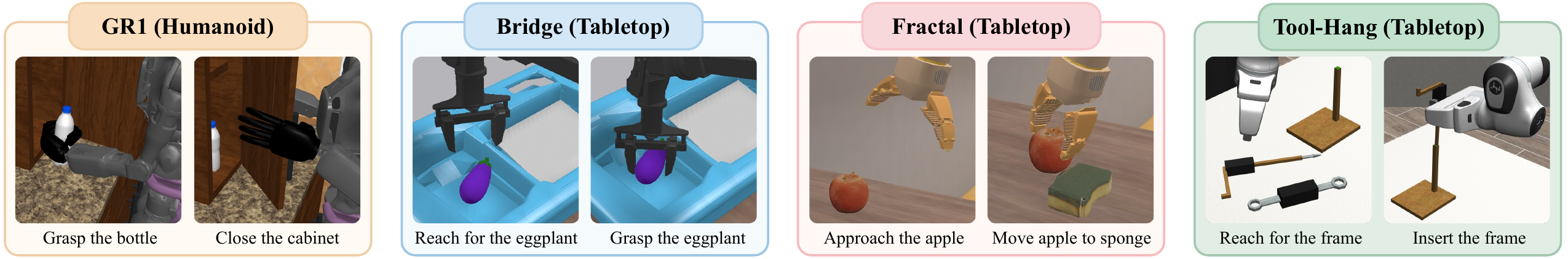}
    \vspace{-1em}
    \caption{\textbf{Example tasks.} We show 4 tasks from RoboCasa-GR1, Bridge, Fractal, and Tool-Hang.}
    \vspace{-1em}
    \label{fig:example-tasks}
\end{figure}

We investigate why regression-based policies perform worse than Flow-Policies from the perspective of statistical modeling: the modeling of prediction residuals and its impact on policy optimization. To conduct this study, we take a pretrained GR00T policy and fine-tune it with the MSE regression loss and the flow-matching loss on three benchmarks: RoboCasa-GR1~\citep{nvidia2025gr00tn1openfoundation}, Bridge~\citep{walke2023bridgedata}, and Fractal~\citep{rt12022arxiv}. We also train a U-Net policy on RoboMimic Tool-Hang~\citep{robomimic2021}. We
collect per-sample predictions and parameter-gradient norms on the training set. Example tasks are visualized in Figure~\ref{fig:example-tasks}. Appendix~\ref{sec:appendix-data-processing} provides details about these benchmarks and our data processing protocols.

\setlength{\intextsep}{\figuretwotextsep}
\begin{wrapfigure}{r}{0.60\textwidth}
  \vspace{-1em}
  \centering  \small
  \captionsetup{font=small,skip=3pt,justification=raggedright,singlelinecheck=false}
  \includegraphics[width=\linewidth]{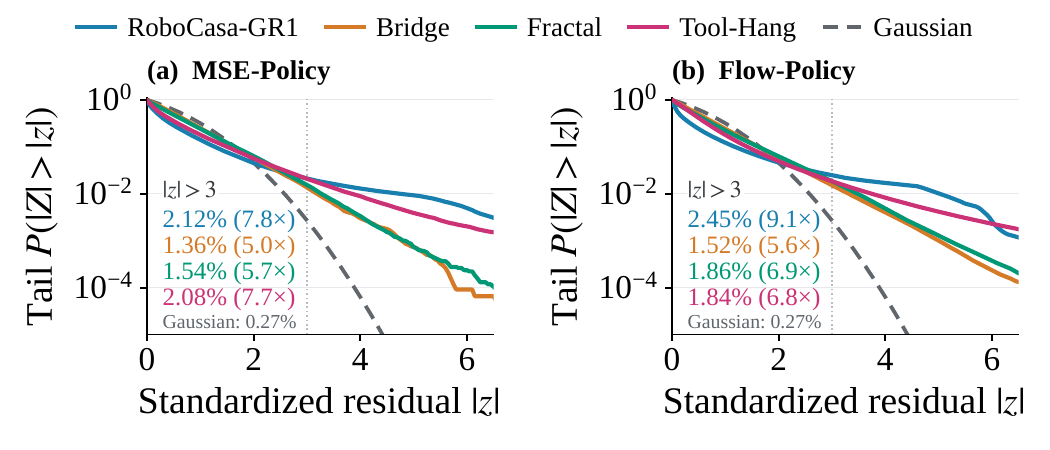}
  \caption{
    \textbf{Heavy-tailed residuals.}
    Both MSE-Policies \textbf{(a)} and Flow-Policies \textbf{(b)} exhibit heavy-tailed action-prediction residuals. Percentages report scalar residuals exceeding three times their coordinate RMS in absolute value.
  }
  \label{fig:heavy-tail}
  \vspace{-1.5\intextsep}
\end{wrapfigure}

\textbf{Action residuals have heavy tails.}
Let $o$ denote the policy inputs and $a\in\mathbb{R}^d$ a demonstrated action chunk. The action residual is $r=\hat a(o)-a$, where $\hat a(o)$ is predicted by an MSE-Policy or sampled from a Flow-Policy using random initial noise. We use $\operatorname{RMS}(r)=\|r\|_2/\sqrt{d}$ for observation-level analyses. To examine tail behavior, Figure~\ref{fig:heavy-tail} pools scalar residuals normalized by their coordinate RMS. Both policy classes exhibit heavier-than-Gaussian tails across the evaluated settings.

From a statistical learning perspective, MSE corresponds to a Gaussian residual model with a fixed scale $\sigma$. Therefore, the results in Figure~\ref{fig:heavy-tail} suggest a real-data mismatch with this model. Next, we illustrate how this will contribute to the performance gap between MSE-Policies and Flow-Policies.

\setlength{\intextsep}{\figuretextsep}

\begin{figure*}[t]
  \centering
  \includegraphics[width=\textwidth]{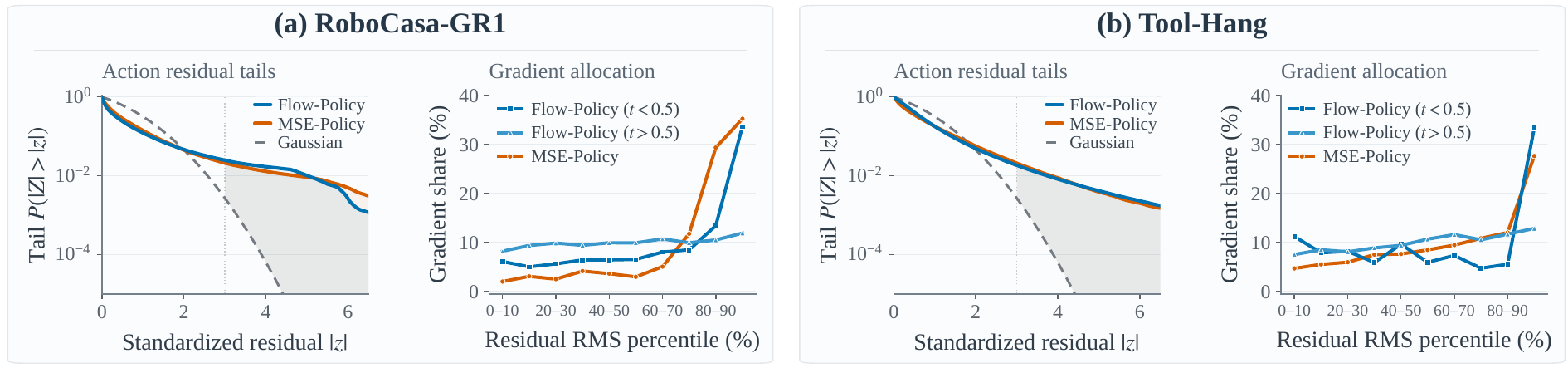}
  \vspace{-1em}
  \caption{
    \textbf{Gradient distribution for different objectives 
    and noise levels.}
    We show the relative gradient contribution (\%) of each
    action-residual decile.
    For Flow, gradient RMS is computed separately for high-noise ($t<0.5$) and low-noise ($t>0.5$) regimes.
    See also the Appendix Figure~\ref{fig:appendix-grad-bin-share}.
  }
  \vspace{-0.5em}
  \label{fig:grad-bin-share}
\end{figure*}

\textbf{Misspecified residual models hurt optimization.}
For MSE-Policies, action samples with large residuals receive disproportionate weight during optimization. Such concentration may leave other action samples underfit. 
To quantify this effect, we measure per-data-point parameter-gradient norms. For each policy, we rank observations by action-residual RMS and divide them into ten equal-sized groups. For Flow-Policies, we compute the average gradient norms over different input noises and quantify gradients in both high-noise ($t<0.5$) and low-noise ($t>0.5$) regimes. Appendix~\ref{sec:appendix-gradient-protocol} provides additional details. Figure~\ref{fig:grad-bin-share} shows the result: MSE and high-noise flow matching concentrate gradients on observations with large action residuals, whereas the low-noise flow-matching objective distributes them more evenly among groups.

\setlength{\intextsep}{\figuretwotextsep}
\begin{wrapfigure}{r}{0.60\textwidth}
  \vspace{-1.5em}
  \centering  \small
  \captionsetup{font=small,skip=3pt,justification=raggedright,singlelinecheck=false}
  \includegraphics[width=\linewidth]{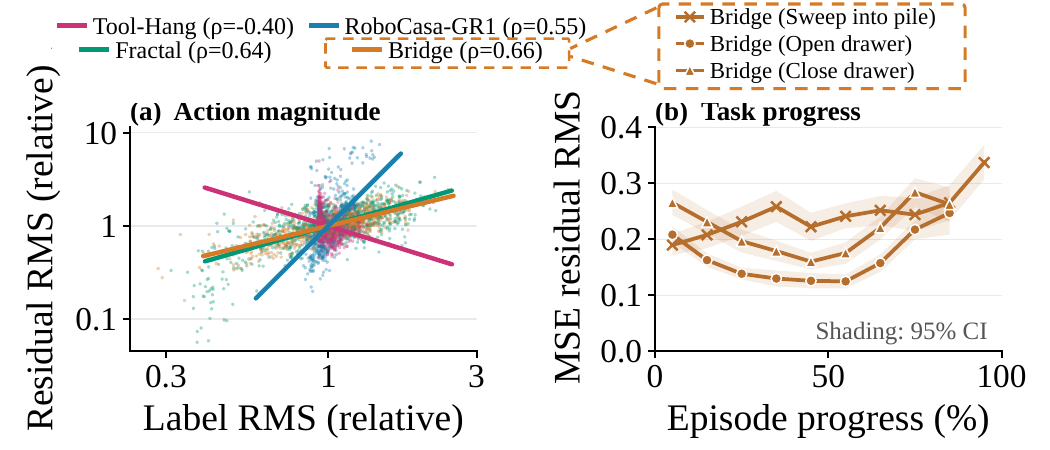}
  \caption{
        \textbf{Input-dependent residual scale.}
      We evaluate a GR00T N1.7 model trained with the MSE regression loss.
      \textbf{(a)} Residual vs. action RMS.
      \textbf{(b)} Residual RMS vs. episode progress on three Bridge tasks. Shaded bands indicate 95\% confidence intervals..
  }
  \label{fig:heterogeneous-scale}
  \vspace{-1em}
\end{wrapfigure}

\textbf{Where do the heavy tails come from?}
We next examine where the heavy-tailed action residuals come from. One potential source is variation in residual scale across observations. Illustrated in Figure~\ref{fig:heterogeneous-scale}, residual RMS varies with both demonstrated action magnitude and episode progress. Even if actions follow a Gaussian distribution per observation, differences in their scales across observations can yield a mixture with heavier-than-Gaussian tails~\citep{andrews1974scale}.

\Needspace{15\baselineskip}
\setlength{\intextsep}{\figuretwotextsep}
\begin{wrapfigure}{r}{0.60\textwidth}
  \vspace{-1.5em}
  \centering  \small
  \captionsetup{font=small,skip=3pt,justification=raggedright,singlelinecheck=false}
  \includegraphics[width=\linewidth]{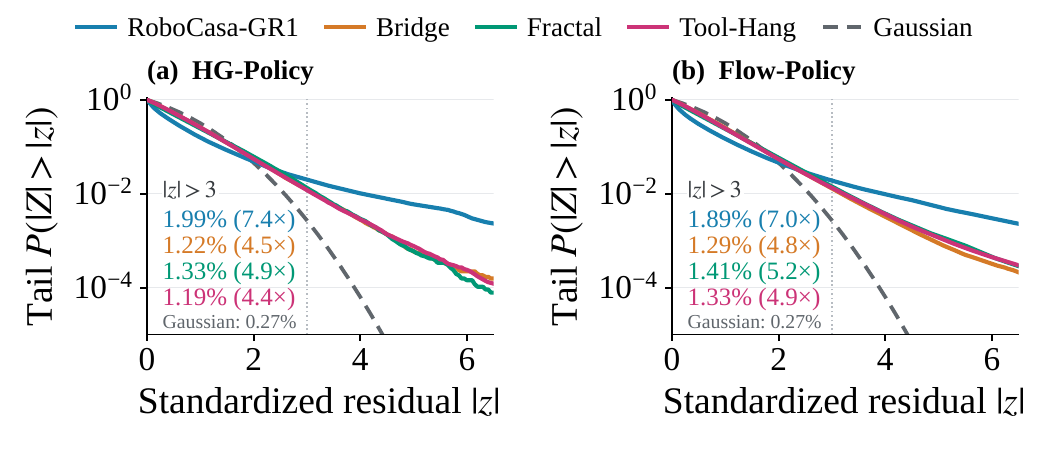}
  \caption{
      \textbf{Heavy tails after normalization.}
      \textbf{(a)} HG-regression residuals normalized by predicted $\sigma(o)$;
      \textbf{(b)} flow-matching residuals normalized by
      per-observation action sampling.
  }
  \label{fig:heavy-tail-normalized}
  \vspace{-1em}
\end{wrapfigure}
Unfortunately, differences in scale do not fully explain the observed heavy tails. To see this, we normalize each action residual by a per-observation scale estimate and re-examine the tails. In particular, we train a model to jointly estimate the mean and variance of actions conditioned on the state using a heteroscedastic Gaussian (HG) objective~\citep{NixPitfall}. We then normalize the residuals by the predicted standard deviations $\sigma(o)$.
For Flow-Policies, we estimate $\sigma(o)$ from multiple actions sampled from the model. Appendix~\ref{sec:appendix-data-processing} provides details. In both cases, normalization makes the tails lighter, but they remain heavier than Gaussian (Figure~\ref{fig:heavy-tail-normalized}).

\textbf{How flow-matching handles heavy-tailed residuals.}
To understand the mechanism of more balanced gradients in flow-matching, we examine its denoising-based training objective. The intuition is that a flow-matching network conditions on a noisy demonstrated action. This additional input allows the training residual to have light tails even when action-prediction residuals are heavy-tailed.

Recall that a flow-matching network takes the policy inputs $o$, a timestep $t$, and a noisy action $a_t=ta+(1-t)\eps$, where $\eps\sim\mathcal{N}(0,I_d)$, and minimizes the squared velocity residual
$v_\theta(a_t,t,o)-(a-\eps)$.
For any fixed $t>0$, consider a na\"ive baseline predictor
$\tilde v(a_t,t,o)=a_t/t$. Its velocity residual is
\[
    \tilde v(a_t,t,o)-(a-\eps)
    = \frac{a_t}{t}-a+\eps
    = \frac{\eps}{t}.
\]
For this baseline model, its network prediction residual is distributed as
$\mathcal{N}(0,I_d/t^2)$, regardless of the overall action distribution. Therefore, the flow-matching objective admits a predictor whose residual scale depends only on the injected noise and the timestep.

Empirically, we found that a well-trained predictor that is conditioned on $o$ also inherits this behavior in the sense that, at low-noise timesteps, the learned predictor has tails no heavier than this reference:
\[
    \Pr\!\left[
        \|v_\theta(a_t,t,o)-(a-\eps)\|_2>\delta
        \,\middle|\, t
    \right]
    \le
    \Pr\!\left[\|\eps\|_2>t\delta\right]
    \quad \text{for sufficiently large } \delta.
\]
Appendix~\ref{sec:appendix-flow-tail-bound} provides empirical evidence on different benchmarks. The denoising objective can tolerate heavy-tailed action residuals because the conditioning of noisy actions allows the velocity residual to be light-tailed. The reference scale $1/t$ decreases as the noise level decreases, which explains the observation of more balanced gradient distributions in low-noise regimes.

\vspace{-0.5em}
\section{Heteroscedastic Student-\texorpdfstring{$t$}{t} Action Regression}
\vspace{-0.5em}
\label{sec:method}

Section~\ref{sec:data-analysis} shows that action residuals have input-dependent scales and heavy tails. MSE-Policies do not account for this modeling and concentrate gradient magnitude on samples with large residuals. We therefore introduce a heteroscedastic Student-$t$ objective for policy learning (HT-Policies). HT-Policies model action residuals with a Student-$t$ distribution whose scale is predicted from the input and whose degrees of freedom $\nu$ control tail heaviness. During optimization, the new HT-Policy likelihood objective downweights gradients from samples with large residuals.

HT-Policies are simple to implement: we augment a standard regression-based action prediction model with an input-dependent residual scale predictor and jointly train both using the following heteroscedastic Student-$t$ (HT) loss:

\textbf{Heteroscedastic Student-$t$ loss.}
Given a data sample $(o, a)$, let $f_\theta(o)\in\mathbb{R}^d$ be the predicted action chunk and $r=f_\theta(o)-a$ the residual. We assume that $r$ follows a multivariate Student-$t$ distribution centered at zero~\citep{lange1989robust}, with fixed degrees of freedom $\nu$ and scale matrix $\sigma_\theta^2(o)$.
The model predicts the action $f_\theta(o)$ and the residual scale $\sigma_\theta(o)$, trained by maximizing the likelihood:
\begin{equation}
\ell_{\mathrm{HT}}(o,a)
=\frac{d}{2}\log\sigma_\theta^2(o)
+\frac{\nu+d}{2}
\log\!\left(1+\frac{\|r\|_2^2}{\nu\sigma_\theta^2(o)}\right),
\label{eq:ht-loss}
\end{equation}
Note that $\sigma_\theta(o)$ is the Student-$t$ scale, rather than the conditional standard deviation in MSE.

To understand how the predicted scale affects each sample's contribution to optimization~\citep{seitzer2022pitfallsheteroscedasticuncertaintyestimation}, we can examine the gradient of the HT loss with respect to the predicted action:
\begin{equation}
\nabla_{f_\theta(o)}\ell_{\mathrm{HT}}
=\frac{r}{\sigma_\theta^2(o)}
\,w\!\left(\frac{\|r\|_2^2}{\sigma_\theta^2(o)}\right),
\qquad
\text{where}~w(s)=\frac{\nu+d}{\nu+s}.
\label{eq:ht-output-gradient}
\end{equation}
This gradient depends on both the predicted scale and the magnitude of the prediction error, relative to that scale. A larger predicted scale reduces the gradient through the factor $1/\sigma_\theta^2(o)$. For a fixed scale, the weight $w(s)$ decreases as the residual grows, limiting the influence of large prediction errors. These gradients then propagate through the action prediction layer to the shared backbone.

\textbf{Selecting the degrees of freedom.}
\label{sec:ht-nu}
The parameter $\nu$ controls how quickly the weight decreases as the residual grows. We set $\nu=cd$, where $d$ is the action dimension, and rewrite the weight as
\begin{equation}
w(s)=\frac{\nu+d}{\nu+s}
=\frac{c+1}{c+s/d},
\qquad
s=\frac{\|r\|_2^2}{\sigma_\theta^2(o)}.
\end{equation}
Here, $s/d$ measures the average squared error relative to the predicted scale across all action dimensions. This reparameterization $\nu=cd$ lets us choose a hyperparameter $c$ independent of the number of action dimensions. A smaller $c$ therefore makes the weight drop more quickly as the error grows. As $c\to\infty$, $w(s)\to 1$, and the HT objective reduces to a heteroscedastic Gaussian objective, retaining scale-based weighting but no longer downweighting samples with large residuals. We study how to choose $c$ empirically in Section~\ref{sec:rq-student-t-ablation}.

\textbf{Fine-tuning pretrained Flow-Policies.}
\label{sec:ht-flow-head}
We can adapt a pretrained flow network into a direct action predictor by setting its action input to zero and its timestep to the pure-noise endpoint. At this input, the optimal flow predictor outputs the conditional mean action, motivating the readout $f_\theta(o)=v_\theta(0,0,o).$ We can then add a scale prediction head and fine-tune the model with the HT loss in Eq.~(\ref{eq:ht-loss}), allowing it to predict an action chunk with a single network evaluation.

To see why this readout is natural, consider the flow path
$x_\tau=(1-\tau)\epsilon+\tau a$, where
$\epsilon\sim\mathcal{N}(0,I_d)$ is sampled independently of the action and policy inputs. The network learns to predict the velocity $a-\epsilon$~\citep{lipman2023flowmatchinggenerativemodeling}. At the pure-noise endpoint $\tau=0$, we have $x_0=\epsilon$, so the optimal predictor under squared error satisfies $
v^*(x,0,o)=\mathbb{E}[a\mid o]-x.$

For models using the reversed path $x_t=(1-t)a+t\epsilon$ with velocity target $\epsilon-a$, the corresponding readout is $f_\theta(o)=-v_\theta(0,1,o)$. Except for changing the prediction target and adding a scale prediction head, we retain each model's original fine-tuning settings in our experiments.
\vspace{-0.5em}
\section{Experiments and Analysis}
\vspace{-0.5em}
\suppressfloats[t] %

In Section~\ref{sec:data-analysis}, we have illustrated how a standard MSE-based--regression policy ({\it MSE-Policy}) fails to model state-dependent residuals and long tails in action learning. This failure leads to imbalanced gradients across state-action pairs with different estimated residuals, which in turn hurts model performance. In the following empirical evaluation, we focus on three questions. First, can the proposed heteroscedastic Student-$t$ policy ({\it HT-Policy}) address state-dependent residuals and long tails, as evaluated by its performance on robot manipulation tasks in both simulation and the real world (Section~\ref{sec:evaluation-scratch})? Second, can we adapt pre-trained vision-language-action (VLA) and world-action (WAM) models to HT-Policy to fully leverage existing community efforts in generalized policy learning? Third, do the data likelihood (residual modeling) and gradient distributions of HT-Policies trained on real-world data support the observations and principles identified in Section~\ref{sec:data-analysis}?

Beyond their performance and general applicability, HT-Policies are a useful policy class for practitioners because they offer greater training and inference efficiency than generative policies, such as flow-matching--based policies ({\it Flow-Policies}), as shown in Section~\ref{sec:evaluation-efficiency}. We conclude this section with ablation studies of different robust-regression objectives and an analysis of the selection of $\nu$.

 Furthermore, HT-Policies offer closed-form likelihood evaluation and efficient action sampling, making them well-suited for reinforcement-learning--based fine-tuning. We evaluate their exploration capability on an example task, Tool-Hang, in Appendix~\ref{sec:appendix-exploration}.
Appendix~\ref{sec:appendix-training} provides additional training and evaluation details, and Appendix~\ref{sec:appendix-loss-settings} provides additional ablation results.

\begin{table*}[!tp]
\centering
\small
\setlength{\tabcolsep}{4.5pt}
\renewcommand{\arraystretch}{1.04}

\begin{minipage}[c]{0.59\textwidth}
\centering
\textbf{(a) RoboMimic}
\vspace{2pt}

\begin{tabular*}{\linewidth}{@{\extracolsep{\fill}}lccc@{}}
\toprule
Backbone & DP & MSE & HT \\
\midrule
U-Net
& 93.8/84.7
& 81.7/72.3
& \textbf{95.3}/\textbf{85.6} \\
Transformer
& \textbf{94.3}/\textbf{84.1}
& 82.6/74.6
& 91.9/82.9 \\
\bottomrule
\label{exp:robomimic}
\end{tabular*}

\end{minipage}
\hfill
\begin{minipage}[c]{0.37\textwidth}
\centering
\textbf{(b) RoboCasa-GR1}
\vspace{2pt}

\begin{tabular*}{\linewidth}{@{\extracolsep{\fill}}ccc@{}}
\toprule
Flow & MSE & HT \\
\midrule
14.0
& 5.8
& \textbf{33.5} \\
\bottomrule
\label{exp:gr00t_scratch}
\end{tabular*}
\end{minipage}
\vspace{-1.5em}
\caption{
Success rates (\%) for policies trained from scratch.
(a) \textbf{RoboMimic.} Best / mean over the last ten evaluated checkpoints~\citep{chi2023diffusionpolicy}, averaged over nine tasks and three seeds per task, with state and image results weighted equally. 
(b) \textbf{RoboCasa-GR1.} A randomly initialized GR00T N1.7 action head trained with the official fine-tuning recipe.
}
\label{tab:rq1-performance}
\vspace{-4pt}
\end{table*}

\FloatBarrier

\vspace{-0.5em}
\subsection{Closing the Performance Gap When Training from Scratch}
\vspace{-0.5em}

\label{sec:evaluation-scratch}

We first investigate whether HT-Policy's modeling of state-dependent and long-tail residuals improves robot manipulation performance when training policies from scratch. We compare HT-Policies with MSE-Policies, diffusion--based policies, and Flow-Policies.

\textbf{Simulation experiment setup.}
On RoboMimic~\citep{robomimic2021}, we train U-Net and Transformer policies in both state-based and image-based settings following Diffusion Policy (DP)~\citep{chi2023diffusionpolicy}.
On RoboCasa-GR1~\citep{nvidia2025gr00tn1openfoundation}, we train action decoders from scratch but using image features
encoded by a frozen VLM backbone from GR00T N1.7 for the vision input.
Appendix~\ref{sec:appendix-gr1-scratch} provides more model and training details.

Table~\ref{tab:rq1-performance} summarizes the results.
On RoboMimic, HT-Policies are on par with generative policies and substantially outperform MSE-based policies.
On RoboCasa-GR1, HT-Policies outperform both MSE-Policies and Flow-Policies.
Overall, HT-Policies are strong direct-regression algorithms for policy training from scratch.
We provide per-task performance breakdowns in Appendix~\ref{sec:appendix-dp} and additional comparisons with MIP policies~\citep{pan2026much} on RoboMimic in Appendix~\ref{sec:appendix-mip}.

\textbf{Real-robot experiment setup.}
Furthermore, we evaluate different policies on three real-world manipulation tasks:
Push-T, Peel-Note, and Insert-T (Figure~\ref{fig:real-robot-eval}).
Training and evaluation details are provided in Appendix~\ref{sec:appendix-real-robot}. 
On the real robot, HT-Policies largely close the performance gap between MSE-based and generative policies on all three tasks. The largest improvements occur on Push-T and Peel-Note, where MSE-Policies fail significantly.

\begin{figure}[!t]
    \centering\small
    \captionsetup{skip=5pt}
    \includegraphics[width=\linewidth]{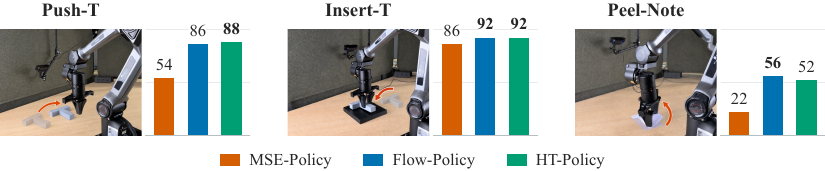}
    \vspace{-1em}
    \caption{\textbf{Real-robot success rates (\%).} We evaluate on 3 tasks: Push-T, Insert-T, and Peel-Note.}
    \label{fig:real-robot-eval}
    \vspace{-0.5em}
\end{figure}

Overall, in both simulation and real-robot tasks, HT-Policies perform on par with generative policies when the action model is trained from scratch. These gains are achieved with a simple feed-forward-only policy, highlighting the importance and effectiveness of residual modeling.

\FloatBarrier
\vspace{-0.5em}
\subsection{Fine-Tuning Pretrained Flow-Policies with HT Regression}
\vspace{-0.5em}
\label{sec:evaluaiton-generality}
\begin{table*}[!tp]
\centering
\small
\setlength{\tabcolsep}{3.5pt}
\begin{tabular*}{\textwidth}{@{\extracolsep{\fill}}llcccc@{}}
\toprule
\multirow{2}{*}{Model} & \multirow{2}{*}{Benchmark}
& \multirow{2}{*}{\makecell{Flow\\(Official)}}
& \multicolumn{3}{c}{Our Evaluations (10 seeds)} \\
\cmidrule(lr){4-6}
& & & Flow & MSE & HT \\
\midrule
$\pi_{0.5}$
& LIBERO (4-suite avg.)
& 96.9 & $\mathbf{97.09\pm0.04}$ & $96.72\pm0.09$ & $96.97\pm0.07$ \\
\midrule
\multirow[t]{3}{*}{GR00T N1.7}
& RoboCasa-GR1
& 44.5 & $42.34\pm0.68$ & $35.55\pm0.67$ & $\mathbf{50.01\pm0.51}$ \\
& SIMPLER / Bridge
& 62.3 & $62.06\pm0.40$ & $58.51\pm0.37$ & $\mathbf{62.31\pm0.65}$ \\
& SIMPLER / Fractal
& 72.5 & $\mathbf{67.83\pm0.39}$ & $60.60\pm0.43$ & $65.25\pm0.44$ \\
\midrule
Cosmos 3
& LIBERO-10
& 95.2 & $95.60\pm0.90$ & $95.10\pm0.26$ & $\mathbf{96.50\pm0.45}$ \\
\bottomrule
\end{tabular*}
\vspace{-3pt}
\caption{
\textbf{Success rate (\%) for fine-tuning from pretrained policies.}
We report the mean success rate and its standard error of a single checkpoint across 10 evaluation seeds.
We include official results for the flow-matching--based fine-tuning for reference.
For $\pi_{0.5}$, we report the average performance in four LIBERO suites.
Bold marks highlight the best-performing model in each row.
}
\label{tab:rq2-performance}
\end{table*}

We next investigate whether HT-Policies can be fine-tuned from pretrained VLA and WAM models to leverage existing advances in generalized policy learning.
Following the procedure described in Section~\ref{sec:method}, the only architectural change is the addition of a randomly initialized residual scale prediction head $\sigma_\theta(o)$. We fine-tune the action predictor using the HT loss while retaining each model's original fine-tuning settings and hyperparameters. We show results on three models:
$\pi_{0.5}$~\citep{intelligence2025pi05visionlanguageactionmodelopenworld}, GR00T N1.7~\citep{nvidia2025gr00tn1openfoundation}, and Cosmos 3~\citep{nvidia2026cosmos3omnimodalworld}, and on three benchmarks: LIBERO~\citep{liu2023liberobenchmarkingknowledgetransfer}, SIMPLER~\citep{li24simpler, walke2023bridgedata, rt12022arxiv}, and RoboCasa-GR1~\citep{nvidia2025gr00tn1openfoundation}. We evaluate each method across 10 evaluation seeds.
These settings span VLA and WAM architectures and cover both fixed-arm and humanoid manipulation. We report the implementation details in Appendix~\ref{sec:appendix-training}.

Shown in Table~\ref{tab:rq2-performance}, HT-Policies achieve higher success rates than MSE-Policies in all five settings, and outperform Flow-Policies in three of them.
These results are obtained using fine-tuning recipes adopted directly from Flow-Policies, rather than specifically tuned for HT-Policies. The only tuned hyperparameter is the learning rate of the newly added scale prediction head, which is trained from scratch.  
We provide per-task performance break-downs in Appendices~\ref{sec:appendix-libero-results}, \ref{sec:appendix-gr1-results}, and~\ref{sec:appendix-cosmos-results}. 

\FloatBarrier
\subsection{Residual Modeling, Gradient Allocation, and Policy Performance}
\label{sec:evaluation-gradient}

The results above show that changing the residual model makes regression-based policies competitive with generative policies.
To further close the theory-implementation gap, we examine the data likelihood and gradient distributions of policies trained on real-world data. This analysis tests whether the principles identified in Section~\ref{sec:data-analysis} regarding residual modeling and gradient allocation explain the observed performance gains.

\begin{figure}[!htb]
    \centering
    \includegraphics[width=\linewidth]{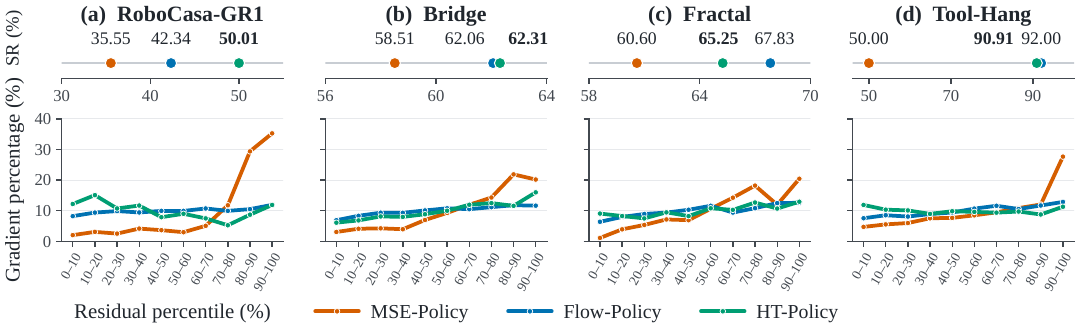}
    \vspace{-0.5em}
    \captionsetup{font=small,skip=5pt,justification=raggedright,singlelinecheck=false}
    \caption{
    \textbf{Residual modeling and optimization.}
    \textbf{Top:} Policy success rates across four settings.
    \textbf{Bottom:} Each action-residual decile's share of summed
    per-sample parameter-gradient RMS under MSE, HT, and low-noise
    Flow ($t>0.5$). Each decile contains approximately $10\%$ of
    the samples.
    }
    \label{fig:gradient-compare}
    \vspace{-1em}
\end{figure}

\textbf{Gradient allocation across observations}
Figure~\ref{fig:gradient-compare} shows that the HT loss produces a more balanced allocation of gradients across observations: large-residual observations no longer dominate the gradients as they do under MSE.
This allocation resembles that of low-noise flow-matching when $t>0.5$.
Appendix~\ref{sec:appendix-gradient-stages} extends this comparison
to different training stages.

\Needspace{19\baselineskip}
\begin{wrapfigure}{r}{0.40\textwidth}
    \centering\small
    \includegraphics[width=\linewidth]{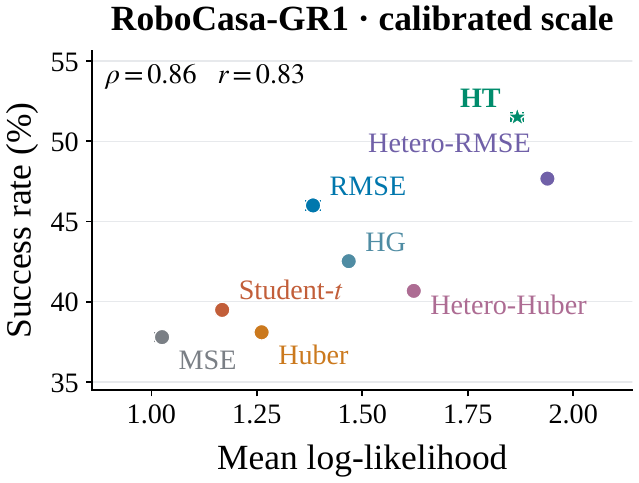}
    \captionsetup{font=small,skip=5pt,justification=raggedright,singlelinecheck=false}
    \caption{
    \textbf{Mean log-likelihood vs. policy success rate.}
    }
    \label{fig:dist-quality-vs-sr}
    \vspace{-0.75\intextsep}
\end{wrapfigure}

\textbf{Residual modeling and policy performance.}
Figure~\ref{fig:dist-quality-vs-sr} examines whether data likelihood under different residual models is correlated with policy performance on RoboCasa-GR1.
Beyond MSE and HT, we include six alternative models, such as models trained with the Huber loss. Across all eight objectives, calibrated mean log-likelihood is strongly correlated with policy success rate (Spearman $\rho=0.86$; Pearson $r=0.83$). HT-Policies achieve substantially higher likelihood than the conventional Gaussian residual model from MSE-based policies, and the highest success rate. Modeling either heavy tails with a Student-$t$ distribution or state-dependent scales with a heteroscedastic Gaussian (HG) also improves likelihood and success over MSE, but neither matches HT, which incorporates both.
Appendix~\ref{sec:likelihood-calibration} provides the evaluation details. Together, these findings support the principles that improved residual modeling and more balanced gradient allocation yield better policy performance.

\FloatBarrier
\vspace{-0.5em}
\subsection{Training and Inference Efficiency}
\label{sec:evaluation-efficiency}
\vspace{-0.5em}

A core advantage of regression policies for practitioners is their training and inference efficiency. We next evaluate whether HT-Policies retain this advantage while achieving strong task performance.

\begin{figure}[!tp]
    \centering
    \small
    \captionsetup{skip=0pt}
    \includegraphics[width=\linewidth]{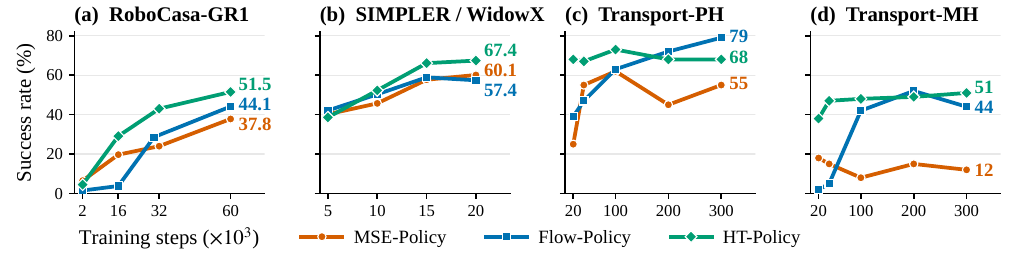}
    \caption{
    \textbf{Training curves for different models on different benchmarks.}
    HT-Policies reach strong task success rates with fewer training steps than Flow-Policies.
    }
    \label{fig:convergence}
\end{figure}

\textbf{Training efficiency.} Shown in Figure~\ref{fig:convergence}, we compare the training progress of HT-Policies, MSE-Policies, and Flow-Policies in two training-from-scratch and two fine-tuning settings.
In all four settings, HT-Policies reach strong task success with fewer training steps than Flow-Policies.

\textbf{Inference efficiency.} HT-Policies require only a single feed-forward pass at inference.
Table~\ref{tab:inference-efficiency} reports batch-size-one inference runtime on an NVIDIA RTX 5090 GPU. HT achieves higher end-to-end throughput than flow on all three models.
Cosmos3-Nano shows the largest whole-model speedup of $17.83\times$, with our HT-Policy replacing the 30-step sampling procedure of Flow-Policies.
Appendix~\ref{sec:appendix-efficiency} provides additional implementation details and results on NVIDIA A800 GPUs.

\begin{table*}[!tp]
\centering\small

\small
\setlength{\tabcolsep}{0pt}
\renewcommand{\arraystretch}{1.08}
\begin{tabular}{@{}
    >{\raggedright\arraybackslash}p{0.19\linewidth}
    *{9}{>{\centering\arraybackslash}p{0.09\linewidth}}
@{}}
\toprule
& \multicolumn{3}{c}{GR00T N1.7}
& \multicolumn{3}{c}{$\pi_{0.5}$}
& \multicolumn{3}{c}{Cosmos3-Nano} \\
\cmidrule(l{4pt}r{4pt}){2-4}
\cmidrule(l{4pt}r{4pt}){5-7}
\cmidrule(l{4pt}r{4pt}){8-10}
Metric
& Flow & HT & Speed-up
& Flow & HT & Speed-up
& Flow & HT & Speed-up \\
\midrule
Head latency\,$\downarrow$
& 29.30 & \textbf{7.92} & $3.70\times$
& 62.50 & \textbf{6.42} & $9.74\times$
& 1266.10 & \textbf{65.96} & $19.19\times$ \\
Total latency\,$\downarrow$
& 46.20 & \textbf{24.52} & $1.88\times$
& 123.58 & \textbf{67.94} & $1.82\times$
& 1271.76 & \textbf{71.34} & $17.83\times$ \\
FPS\,$\uparrow$
& 21.65 & \textbf{40.78} & $1.88\times$
& 8.09 & \textbf{14.72} & $1.82\times$
& 0.79 & \textbf{14.02} & $17.83\times$ \\
\bottomrule
\end{tabular}
\captionsetup{skip=5pt}
\caption{\textbf{Inference efficiency benchmarking.} We evaluate all policies on a single NVIDIA RTX 5090 GPU with batch size 1. Latencies are in milliseconds.}
\label{tab:inference-efficiency}
\vspace{-1em}
\end{table*}

\FloatBarrier
\vspace{-0.5em}
\subsection{Ablation Studies}
\label{sec:ablation}
\vspace{-0.5em}

\Needspace{18\baselineskip}
\begin{wrapfigure}{r}{0.65\textwidth}
    \centering
    \includegraphics[height=0.417\linewidth]{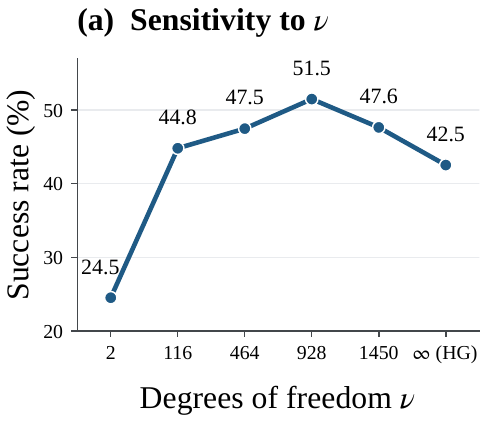}\hfill
    \includegraphics[height=0.417\linewidth]{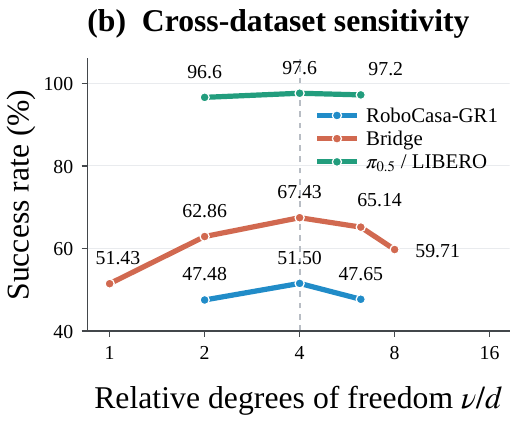}
    \captionsetup{font=small,skip=5pt,justification=raggedright,singlelinecheck=false}
    \caption{
    \textbf{Selection of $\nu$.}
    \textbf{(a)} Success rate vs. $\nu$ on RoboCasa-GR1 and
    \textbf{(b)} Success rate vs. $c=\nu/d$ on three datasets, where $d$ is the action-chunk dimension.
    $\nu\to\infty$ gives heteroscedastic Gaussian.
    }
    \label{fig:nu-sensitivity}
    \vspace{-0.85\intextsep}
\end{wrapfigure}

\textbf{Ablation on the degrees of freedom $\nu$.}
\label{sec:rq-student-t-ablation}
In practice, we select $\nu$ in proportion to the action-chunk dimension $d$, parameterizing it as $\nu=cd$. Figure~\ref{fig:nu-sensitivity}a shows that RoboCasa-GR1 performance peaks at an intermediate value of $\nu$, favoring moderate downweighting of large residuals over either strong downweighting or the Gaussian limit.
To evaluate this dimension-dependent choice across settings, Figure~\ref{fig:nu-sensitivity}b compares performance as a function of $c=\nu/d$. RoboCasa-GR1, Bridge, and LIBERO all achieve their highest tested success rates at $c=4$, supporting $\nu=4d$ as a practical default across these settings.
Appendix~\ref{sec:appendix-nu-settings} provides evaluation details.

\textbf{Comparison with alternative training objectives.}
Figure~\ref{fig:dist-quality-vs-sr} summarizes the comparison among HT-Policies and alternative robust-regression objectives by calibrated likelihood and performance; Appendix~\ref{sec:residual-model-ablation} defines these objectives (Table~\ref{tab:app-loss-definitions}) and provides detailed results (Table~\ref{tab:robust-loss-ablation}). Overall, HT-Policies achieve the best performance.

\vspace{-0.5em}
\section{Conclusion}
\vspace{-0.5em}
In this paper, we revisit the performance gap between direct action-regression policies and generative policies through the lens of statistical modeling of prediction residuals and their effects on optimization. Contrary to the common hypothesis that generative policies perform better because they better model multimodal action distributions, our analysis suggests that the empirical performance gap is better explained by differences in how these methods handle network prediction residuals. We show that real-world demonstrations exhibit input-dependent residual scales and heavy tails. The fixed-scale residual model implicit in MSE regression leads to gradient imbalance, whereas the denoising objectives in generative policies better mitigate this. These findings motivate HT-Policies, a class of neural policies that are performant without iterative computation, can be trained from scratch or fine-tuned from pretrained models, and offer improved training and inference efficiency.

\textbf{Limitations.} So far, our analysis has focused on network fitting and prediction errors, without directly connecting them to control failures. In particular, it does not explain why small errors at a few states can catastrophically reduce overall success rate. Incorporating control-theoretic analysis is an important next step, such as in \citet{tu2022stability}. Second, HT-Policies can not explicitly represent multiple action modes. Our contribution offers an explanation for the observed gap between regression and generative policies and an efficient alternative for tasks with limited multimodality, rather than a universal solution to policy learning. Finally, although the appendix provides preliminary evidence that HT-Policies offer a promising starting point for RL due to their closed-form likelihood and action sampling, a systematic evaluation remains future work.

\bibliography{iclr2027_conference}
\bibliographystyle{iclr2027_conference}

\clearpage
\appendix
\section{Appendix}
\begingroup
\newlength{\appendixtaskwidth}
\makeatletter
\renewcommand{\subsection}{\@startsection{subsection}{2}{\z@}
  {-1.3ex plus -.4ex minus -.2ex}{.6ex plus .2ex}
  {\normalfont\normalsize\scshape\raggedright}}
\renewcommand{\subsubsection}{\@startsection{subsubsection}{3}{\z@}
  {-1.0ex plus -.3ex minus -.2ex}{.4ex plus .2ex}
  {\normalfont\normalsize\scshape\raggedright}}
\renewcommand{\paragraph}{\@startsection{paragraph}{4}{\z@}
  {0pt}{-.5em}{\normalfont\normalsize\bfseries}}
\makeatother
\captionsetup{font=small,labelfont=normalfont,skip=5pt,
  justification=justified,singlelinecheck=false}
\clubpenalty=10000
\widowpenalty=10000
\renewcommand{\topfraction}{0.95}
\renewcommand{\bottomfraction}{0.85}
\renewcommand{\textfraction}{0.05}
\renewcommand{\floatpagefraction}{0.8}
\setcounter{topnumber}{4}
\setcounter{bottomnumber}{3}
\setcounter{totalnumber}{6}

\subsection{Residual and Gradient Diagnostics}
\label{sec:appendix-diagnostics}

\subsubsection{Residual Definitions and Sampling}
\label{sec:appendix-data-processing}

We analyze residuals and gradients on the training demonstrations,
as in Section~\ref{sec:data-analysis}. The calibration splits below
hold out observations only for scale estimation.

\paragraph{Policies and samples.}
Figures~\ref{fig:heavy-tail} and~\ref{fig:heavy-tail-normalized}
use GR00T MSE-Policies, HG-Policies, and Flow-Policies trained for 60k
updates on RoboCasa-GR1 and 20k updates on Bridge and Fractal.
We select ten episodes from each
of the 24 GR1 tasks. For Bridge and Fractal, we select the
40 most frequent instruction groups and take ten episodes
per group. Each episode contributes four evenly spaced,
complete eight-step action chunks. Every method uses the
same selected observations and demonstrated actions.
These groups describe the training demonstrations and are
separate from the seven and six SIMPLER evaluation tasks.
For Tool-Hang PH, we evaluate state-based MSE-Policies, HG-Policies,
and Flow-Policies with Chi-UNet. All three methods share
eight evenly spaced, complete 16-step chunks from each of the
200 demonstrations.

\begin{table}[htbp]
\centering
\caption{\textbf{Residual-tail samples.} Bridge, Fractal, and Tool-Hang exclude the
gripper channel from the plotted residuals. Each Flow-Policy contributes
64 independently evaluated action draws per observation.}
\label{tab:app-tail-samples}
\small
\renewcommand{\arraystretch}{1.1}
\begin{tabularx}{\linewidth}{@{}Xrrrr@{}}
\toprule
Dataset & Episodes & Observations & $d$ & MSE/HG coordinates \\
\midrule
RoboCasa-GR1 & 240 & 960 & 232 & 222,720 \\
Bridge & 400 & 1,600 & 48 & 76,800 \\
Fractal & 400 & 1,600 & 48 & 76,800 \\
Tool-Hang & 200 & 1,600 & 144 & 230,400 \\
\bottomrule
\end{tabularx}
\end{table}

\paragraph{Residual definition.}
We use the action residual $r=\hat a(o)-a$ defined in the main text,
computed in normalized action space without augmentation or dropout.
The chunk magnitude $\operatorname{RMS}(r)=\|r\|_2/\sqrt{d}$ is used
for observation-level analyses and gradient grouping. The tail plots
pool scalar residual coordinates, each normalized by its coordinate
RMS as specified below.
GR1 retains its 29 valid action channels. Bridge and Fractal
retain six continuous channels and omit the gripper from the
tail plots; the HG-Policy's predicted scale still uses its original
training mask, including the gripper.
Tool-Hang retains nine continuous channels with 6D rotation encoding.
For MSE-Policies and HG-Policies, residuals subtract the demonstration
from the predicted mean. For Flow-Policies, residuals subtract it from
each sampled action, without averaging the generated actions first.

\paragraph{Fixed coordinate scales.}
For GR00T, the ten episodes within each task or instruction group are
divided into five folds of two episodes.
Tool-Hang uses five folds of 40 episodes each.
For each evaluated fold, the other four folds estimate a fixed
RMS for each chunk coordinate $j$:
\begin{equation}
    q_j=\sqrt{\E_{\mathrm{cal}}[u_j^2]},
    \qquad z_{ij}=u_{ij}/q_j.
    \label{eq:app-coordinate-calibration}
\end{equation}
Here $u_{ij}$ is either the raw residual or the residual after
division by the observation-dependent scale described below.
For Flow-Policies, expectations also average over the designated action draws.
There is no mean subtraction, residual trimming, or subsequent
rescaling using the evaluated fold.
All valid scalar coordinates are pooled to estimate $\Pr(|z|>x)$.
The percentages in Figure~\ref{fig:heavy-tail} report $\Pr(|z|>3)$:
the fraction of scalar residuals exceeding three times their coordinate
RMS in absolute value. The two-sided standard Gaussian reference is
approximately $0.27\%$ at this threshold.

\paragraph{Observation-dependent scales.}
For each HG-Policy, $u_{ij}=r_{ij}/\sigma_i$ uses its predicted scalar scale.
For Flow-Policies, we draw 128 independent samples per observation
with unit Gaussian initial noise, using the native four-step
sampler for GR00T and nine Euler steps for Tool-Hang.
The first 64 estimate spread and the remaining 64 enter
the tail curves. Let $v_{ij}$ be the unbiased sample variance
of coordinate $j$ in the first bank. The calibration folds
define $b_j=\E_{\mathrm{cal}}[v_{ij}]$, and the local spread is
\begin{equation}
    s_i=\sqrt{\frac{1}{d}\sum_j\frac{v_{ij}}{b_j}}.
\end{equation}
The own-spread curve uses $u_{ij}=(A_{ij}-a_{ij})/s_i$,
followed by Eq.~(\ref{eq:app-coordinate-calibration}).
The separate sampling banks keep an extreme evaluated draw
from increasing its own normalization scale.

\subsubsection{Per-Observation Parameter Gradients}
\label{sec:appendix-gradient-protocol}

\begin{figure}[!htbp]
    \centering
    \includegraphics[width=\linewidth]{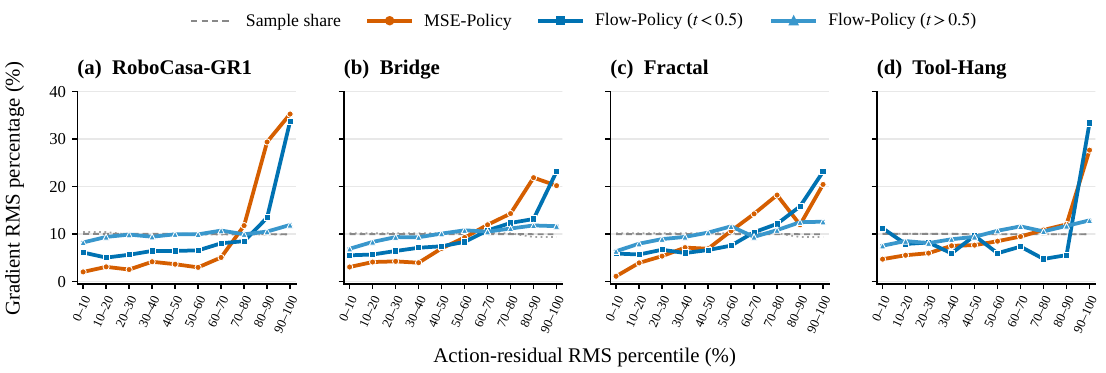}
    \caption{
    \textbf{Gradient imbalance across objectives and noise levels.}
    Curves show the relative gradient contribution (\%) of each
    action-residual decile.
    For flow matching, gradient RMS is averaged over noise draws and timesteps
    separately for high noise ($t<0.5$) and low noise ($t>0.5$).
    }
    \label{fig:appendix-grad-bin-share}
\end{figure}

\paragraph{Parameter scope and sample counts.}
Figures~\ref{fig:grad-bin-share} and~\ref{fig:gradient-compare}
use gradients of individual observation losses before the
optimizer, clipping, or weight decay.
For GR00T, the parameter set contains the trainable action-head
internals, excluding the final action and scale decoders and
the frozen VLM.
For Tool-Hang, it contains the native observation encoder and
shared Chi-UNet trunk, excluding the output heads.
Gradients of the HT loss combine the mean and scale paths before the norm
is computed, retaining their cross term.
Table~\ref{tab:app-gradient-samples} gives the number of observations
evaluated for each method.

\begin{table}[htbp]
\centering
\caption{\textbf{Observation counts for the parameter-gradient diagnostics.}
GR00T is trained for 60k updates on GR1 and 20k updates on Bridge
and Fractal. Tool-Hang uses state observations and Chi-UNet.}
\label{tab:app-gradient-samples}
\small
\renewcommand{\arraystretch}{1.1}
\begin{tabularx}{\linewidth}{@{}Xrrr@{}}
\toprule
Setting & MSE & Flow & HT \\
\midrule
RoboCasa-GR1 & 1,536 & 1,536 & 1,536 \\
Bridge & 2,048 & 2,048 & 2,048 \\
Fractal & 2,048 & 2,048 & 2,048 \\
Tool-Hang & 4,096 & 4,096 & 4,096 \\
\bottomrule
\end{tabularx}
\end{table}

\paragraph{Residual ranking and gradient shares.}
Each policy's observations are ranked by its own action-chunk
residual RMS and divided into ten groups of nearly equal size.
All valid action channels, including grippers, enter this ranking.
MSE-Policies and HT-Policies use their predicted means. Flow-Policies
use the complete native sampler starting from zero initial noise.
For raw parameter gradient $g_{ikp}$ of observation $i$, draw
$k$, and parameter $p$, the reported contribution is
\begin{equation}
    m_i=\frac{1}{K}\sum_{k=1}^{K}
         \sqrt{\frac{1}{P}\sum_{p=1}^{P}g_{ikp}^2},
    \qquad
    G_b=\frac{\sum_{i\in b}m_i}{\sum_i m_i}.
    \label{eq:app-gradient-share}
\end{equation}
The figures display $100G_b$. Norms are computed before
averaging draws, and observations are summed only afterwards.
This is not the norm of an accumulated batch gradient.

\paragraph{Flow-matching noise and timestep groups.}
Flow-matching gradients use the native velocity MSE at 16 equally weighted
midpoint quantiles of the training timestep distribution,
$t_k=Q((k-\tfrac12)/16)$ for $k=1,\ldots,16$, with independent
Gaussian action noise for each observation and draw.
Each point represents a quantile interval with probability $1/16$.
The high- and low-noise curves retain points with $t_k<0.5$ and
$t_k>0.5$, respectively, before applying Eq.~(\ref{eq:app-gradient-share}).
Table~\ref{tab:app-flow-timestep-groups} summarizes the resulting groups.
The thresholds apply to the representative points, so the represented
intervals need not meet at exactly $0.5$.
These two flow-matching curves share the same observations and residual
bins. Gradients are measured with training-mode dropout and augmentation.

\begin{table}[htbp]
\centering
\caption{\textbf{Flow-matching timestep groups for gradient diagnostics.}
Ranges span the full quantile intervals represented by each group;
percentages denote their training probability mass.}
\label{tab:app-flow-timestep-groups}
\small
\setlength{\tabcolsep}{4pt}
\renewcommand{\arraystretch}{1.1}
\begin{tabularx}{\linewidth}{@{}Xccc@{}}
\toprule
Dataset & Quantile function $Q(u)$ & High noise ($t<0.5$) & Low noise ($t>0.5$) \\
\midrule
\makecell[l]{RoboCasa-GR1,\\Bridge, Fractal}
& $0.999[1-(1-u)^{2/3}]$
& \makecell{Points 1--10 ($62.5\%$)\\$[0,\,0.4795)$}
& \makecell{Points 11--16 ($37.5\%$)\\$[0.4795,\,0.999]$} \\
\addlinespace[2pt]
Tool-Hang & $u$
& \makecell{Points 1--8 ($50\%$)\\$[0,\,0.5)$}
& \makecell{Points 9--16 ($50\%$)\\$[0.5,\,1]$} \\
\bottomrule
\end{tabularx}
\end{table}

\FloatBarrier
\Needspace{10\baselineskip}
\subsubsection{Gaussian Tail Bounds for Flow-Matching Velocity Residuals}
\label{sec:appendix-flow-tail-bound}

We examine the tail-bound hypothesis in Section~\ref{sec:tail-tolerance}
using the velocity residual
$r_\theta=v_\theta(a_t,t,o)-(a-\eps)$.
At six fixed low-noise timesteps ($t>0.5$), we compare the empirical
tail probabilities of $t\|r_\theta\|_2/\sqrt{d}$ with the Gaussian
norm reference $\|\eps\|_2/\sqrt{d}\sim\chi_d/\sqrt{d}$.
The largest empirical tail probability across these timesteps
remains below the reference over the plotted range across datasets
(Figure~\ref{fig:flow-tail-bound-datasets}) and training stages on
RoboCasa-GR1 (Figure~\ref{fig:flow-tail-bound-stages}), supporting
the hypothesis in the evaluated low-noise settings.

\begin{figure}[!htbp]
    \centering
    \includegraphics[width=\linewidth]{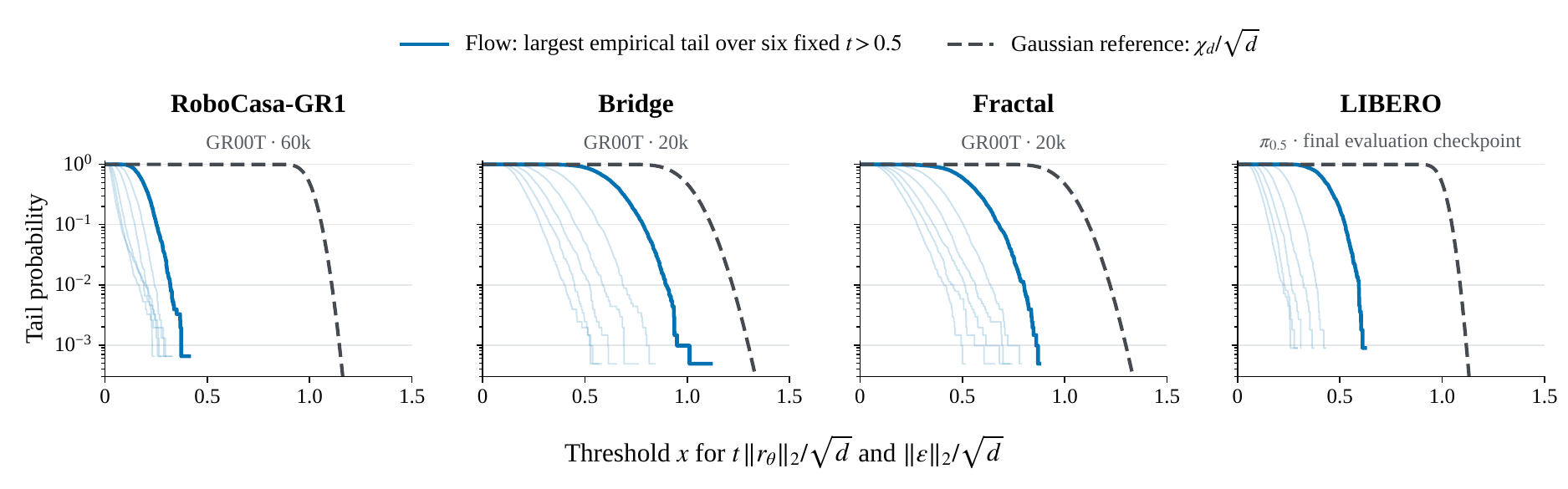}
    \caption{\textbf{Gaussian tail-bound comparison across datasets.}
    GR00T uses 60k updates on RoboCasa-GR1 and 20k on Bridge and
    Fractal; LIBERO results use $\pi_{0.5}$.
    Light curves show individual timesteps, the thick blue curve their
    pointwise maximum, and the dashed curve the Gaussian norm reference.}
    \label{fig:flow-tail-bound-datasets}
\end{figure}

\begin{figure}[!htbp]
    \centering
    \includegraphics[width=\linewidth]{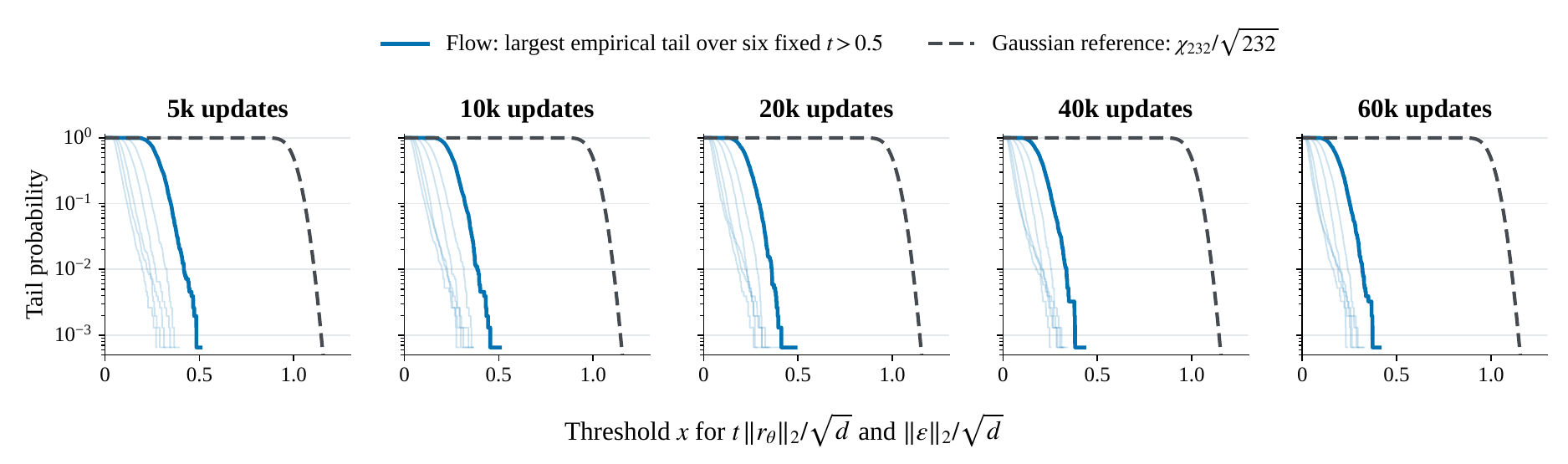}
    \caption{\textbf{Gaussian tail-bound comparison across training stages.}
    GR00T on RoboCasa-GR1 after 5k, 10k, 20k, 40k, and 60k updates,
    with $d=232$. Curves follow the same convention as
    Figure~\ref{fig:flow-tail-bound-datasets}.}
    \label{fig:flow-tail-bound-stages}
\end{figure}
\FloatBarrier

\Needspace{8\baselineskip}
\subsubsection{Gradient Allocation across Training Stages}
\label{sec:appendix-gradient-stages}

We examine whether the gradient patterns in
Section~\ref{sec:evaluation-gradient} are already present early
in training. Figure~\ref{fig:gr1-stage-gradient-sr} compares MSE-Policies,
Flow-Policies, and HT-Policies on RoboCasa-GR1 after 5k, 10k, 20k, 40k, and 60k
optimizer updates. At each stage, it shows how the summed
per-sample gradient RMS is distributed across action-residual
deciles, together with the corresponding policy success rate.
We follow the gradient-diagnostic protocol in
Appendix~\ref{sec:appendix-gradient-protocol}, including low-noise
flow-matching timesteps ($t>0.5$). These stage-wise results are
reported separately from the main benchmark comparison.

\begin{figure}[!htbp]
    \centering
    \includegraphics[width=\linewidth]{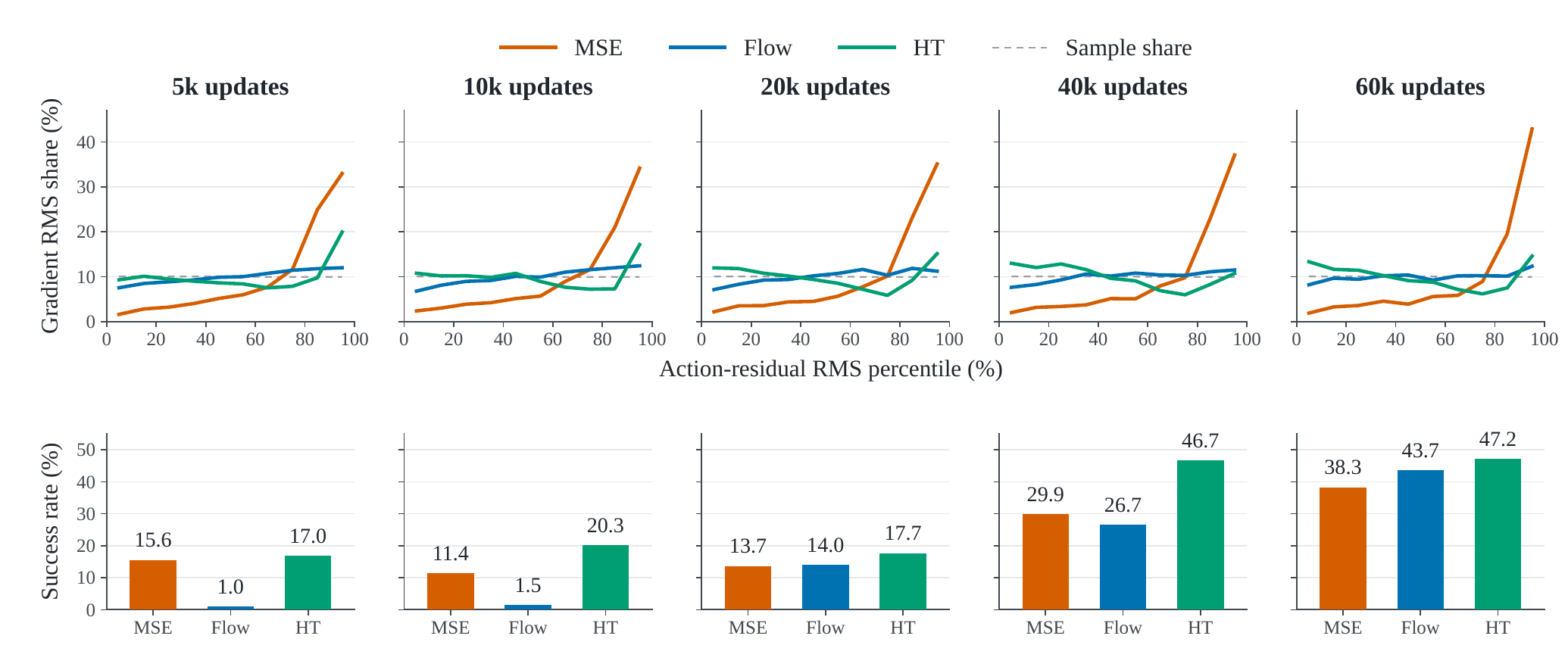}
    \caption{\textbf{Gradient allocation and policy success across
    training stages on RoboCasa-GR1.}
    Columns show results after 5k, 10k, 20k, 40k, and 60k updates.
    \textbf{Top:} Each action-residual decile's share of summed
    per-sample parameter-gradient RMS, with flow-matching gradients evaluated
    at low-noise timesteps ($t>0.5$). The dashed line marks the
    $10\%$ sample share of each decile.
    \textbf{Bottom:} Success rates of MSE-Policies, Flow-Policies, and
    HT-Policies at the corresponding training stages; HT-Policies have the
    highest success rate at all five stages.}
    \label{fig:gr1-stage-gradient-sr}
\end{figure}

The MSE loss assigns a disproportionate gradient share to the largest
residuals at every evaluated stage, and its highest-residual
decile becomes more dominant as training progresses. The flow-matching
and HT losses distribute gradient magnitude more evenly across residual
deciles throughout the comparison. In particular, the HT loss already
reduces the concentration of gradients in the residual tail at
5k updates, well before HT-Policies reach their strongest performance at
40k and 60k. The reduced tail influence is therefore visible
early and persists across the evaluated training stages.

\Needspace{10\baselineskip}
\subsection{Action Averaging in Flow-Policies}
\label{sec:appendix-flow-action-averaging}

We examine whether averaging actions sampled from a trained Flow-Policy affects task success. We keep each policy fixed and compare
its native single-sample output ($K=1$) with the mean of $K=16$
complete action chunks. At each replanning step, we condition all
samples on the current observation $o$ and use distinct initial noise:
\begin{equation}
    A^{(k)}=\operatorname{Flow}_{\theta}(o,\epsilon_k),
    \qquad
    \bar A=\frac{1}{16}\sum_{k=1}^{16}A^{(k)}.
    \label{eq:flow-action-averaging}
\end{equation}
Each sample uses the full native sampler: 4 steps for GR00T N1.7,
10 for $\pi_{0.5}$, and 30 for Cosmos3-Nano. We average the final
chunks in normalized action space, including gripper and
finger channels, before applying the original action
postprocessing and execution. Both conditions use the same trained
policy, observation preprocessing, and action execution length.

\paragraph{Evaluation.}
We evaluate all 24 RoboCasa-GR1 tasks, all 40 tasks in the four
LIBERO suites for $\pi_{0.5}$, and all 10 LIBERO-10 tasks for
Cosmos3-Nano. The requested rollout counts per task and condition
are 20, 100, and 10, respectively; each query executes 8, 10,
and 16 actions. In particular, $\pi_{0.5}$ executes only the first
10 actions of its 50-step prediction. Cosmos3-Nano retains joint
video/action denoising for every sample and averages only the final
actions. Its $K=16$ mean includes the $K=1$ sample and 15 additional
noise initializations.

Each setting uses one evaluation run per condition, with $K=1$
remeasured for this comparison. These results are separate from the
10-seed means in Table~\ref{tab:rq2-performance}. RoboCasa-GR1 has
480 and 483 completed rollouts for $K=1$ and $K=16$, respectively.
We compute each task's success rate using its actual rollout count
and average task rates equally. For $\pi_{0.5}$, we average task
rates within each suite and then average the four suites.

\begin{table}[!htbp]
\centering
\caption{\textbf{Action averaging in fixed Flow-Policies.}
Success rates (\%) from one evaluation run per condition.
$\Delta$ is the change in percentage points, computed before rounding.}
\label{tab:flow-action-averaging}
\small
\setlength{\tabcolsep}{6pt}
\renewcommand{\arraystretch}{1.1}
\begin{tabular*}{\linewidth}{@{\extracolsep{\fill}}lrrr@{}}
\toprule
Model / benchmark & $K=1$ & $K=16$ mean & $\Delta$ (pp) \\
\midrule
GR00T N1.7 / RoboCasa-GR1 & 40.63 & 43.23 & $+2.61$ \\
$\pi_{0.5}$ / LIBERO & 97.250 & 97.325 & $+0.075$ \\
Cosmos3-Nano / LIBERO-10 & 96.00 & 98.00 & $+2.00$ \\
\bottomrule
\end{tabular*}
\end{table}

Table~\ref{tab:flow-action-averaging} shows no decrease in
benchmark-average success under action averaging in these three
evaluations, although individual tasks can improve or decline.
These results suggest that averaging sampled action chunks need not
harm performance in the tested policies; the single-run comparisons
do not establish a consistent improvement across evaluation seeds.
Tables~\ref{tab:flow-averaging-gr1}--\ref{tab:flow-averaging-cosmos}
provide all task-level results.

\begin{table}[!htbp]
\centering
\small
\setlength{\tabcolsep}{5pt}
\renewcommand{\arraystretch}{1.08}
\caption{\textbf{Action averaging on RoboCasa-GR1.}
Entries give successes / completed rollouts, with success rates (\%) in
parentheses. Task names follow Table~\ref{tab:appendix-gr1-results}.
The mean weights all 24 tasks equally; $\Delta$ is computed before rounding.}
\label{tab:flow-averaging-gr1}
\begin{tabular}{@{}lrrr@{}}
\toprule
Task & $K=1$ & $K=16$ mean & $\Delta$ (pp) \\
\midrule
PnPBottle $\to$ CabinetClose & $8/20$ (40.00) & $12/20$ (60.00) & $+20.00$ \\
PnPCan $\to$ DrawerClose & $2/20$ (10.00) & $6/20$ (30.00) & $+20.00$ \\
PnPCup $\to$ DrawerClose & $0/20$ (0.00) & $4/20$ (20.00) & $+20.00$ \\
PnPMilk $\to$ MicrowaveClose & $3/20$ (15.00) & $1/20$ (5.00) & $-10.00$ \\
PnPPotato $\to$ MicrowaveClose & $4/20$ (20.00) & $1/20$ (5.00) & $-15.00$ \\
PnPWine $\to$ CabinetClose & $3/20$ (15.00) & $3/21$ (14.29) & $-0.71$ \\
\addlinespace[2pt]
Cuttingboard $\to$ Basket & $14/20$ (70.00) & $15/20$ (75.00) & $+5.00$ \\
Cuttingboard $\to$ Cardboardbox & $8/20$ (40.00) & $6/20$ (30.00) & $-10.00$ \\
Cuttingboard $\to$ Pan & $15/20$ (75.00) & $14/20$ (70.00) & $-5.00$ \\
Cuttingboard $\to$ Pot & $14/20$ (70.00) & $13/20$ (65.00) & $-5.00$ \\
Cuttingboard $\to$ Tieredbasket & $6/20$ (30.00) & $10/20$ (50.00) & $+20.00$ \\
\addlinespace[2pt]
Placemat $\to$ Basket & $12/20$ (60.00) & $14/20$ (70.00) & $+10.00$ \\
Placemat $\to$ Bowl & $3/20$ (15.00) & $7/20$ (35.00) & $+20.00$ \\
Placemat $\to$ Plate & $8/20$ (40.00) & $9/20$ (45.00) & $+5.00$ \\
Placemat $\to$ Tieredshelf & $1/20$ (5.00) & $3/20$ (15.00) & $+10.00$ \\
\addlinespace[2pt]
Plate $\to$ Bowl & $8/20$ (40.00) & $14/20$ (70.00) & $+30.00$ \\
Plate $\to$ Cardboardbox & $9/20$ (45.00) & $4/21$ (19.05) & $-25.95$ \\
Plate $\to$ Pan & $9/20$ (45.00) & $6/20$ (30.00) & $-15.00$ \\
Plate $\to$ Plate & $16/20$ (80.00) & $13/20$ (65.00) & $-15.00$ \\
\addlinespace[2pt]
Tray $\to$ Cardboardbox & $10/20$ (50.00) & $14/20$ (70.00) & $+20.00$ \\
Tray $\to$ Plate & $13/20$ (65.00) & $15/20$ (75.00) & $+10.00$ \\
Tray $\to$ Pot & $15/20$ (75.00) & $13/20$ (65.00) & $-10.00$ \\
Tray $\to$ Tieredbasket & $10/20$ (50.00) & $8/20$ (40.00) & $-10.00$ \\
Tray $\to$ Tieredshelf & $4/20$ (20.00) & $3/21$ (14.29) & $-5.71$ \\
\midrule
Mean success (\%) & 40.63 & 43.23 & $+2.61$ \\
\bottomrule
\end{tabular}
\end{table}

\begin{table}[!htbp]
\centering
\small
\setlength{\tabcolsep}{5pt}
\renewcommand{\arraystretch}{1.08}
\setlength{\tabcolsep}{4pt}
\caption{\textbf{Action averaging for $\pi_{0.5}$ on all four LIBERO suites.}
Entries give successes / completed rollouts. Task IDs are listed in
Tables~\ref{tab:app-libero-tasks} and~\ref{tab:app-libero-goal-long-tasks}.}
\label{tab:flow-averaging-pi05}
\begin{tabular}{@{}rrrrrrrrr@{}}
\toprule
& \multicolumn{2}{c}{LIBERO-10} & \multicolumn{2}{c}{LIBERO-Goal}
& \multicolumn{2}{c}{LIBERO-Object} & \multicolumn{2}{c}{LIBERO-Spatial} \\
\cmidrule(lr){2-3}\cmidrule(lr){4-5}\cmidrule(lr){6-7}\cmidrule(l){8-9}
ID & $K=1$ & $K=16$ & $K=1$ & $K=16$ & $K=1$ & $K=16$ & $K=1$ & $K=16$ \\
\midrule
0 & $96/100$ & $93/100$ & $99/100$ & $99/100$ & $100/100$ & $100/100$ & $100/100$ & $100/100$ \\
1 & $98/100$ & $98/100$ & $98/100$ & $100/100$ & $100/100$ & $100/100$ & $100/100$ & $99/100$ \\
2 & $99/100$ & $96/100$ & $89/100$ & $91/100$ & $100/100$ & $100/100$ & $100/100$ & $100/100$ \\
3 & $100/100$ & $100/100$ & $94/100$ & $94/100$ & $97/100$ & $98/100$ & $97/100$ & $100/100$ \\
4 & $99/100$ & $98/100$ & $100/100$ & $100/100$ & $100/100$ & $99/100$ & $98/100$ & $98/100$ \\
5 & $100/100$ & $100/100$ & $99/100$ & $99/100$ & $100/100$ & $100/100$ & $80/100$ & $84/100$ \\
6 & $95/100$ & $95/100$ & $96/100$ & $95/100$ & $100/100$ & $100/100$ & $100/100$ & $99/100$ \\
7 & $99/100$ & $98/100$ & $100/100$ & $100/100$ & $99/100$ & $100/100$ & $99/100$ & $96/100$ \\
8 & $72/100$ & $79/100$ & $100/100$ & $100/100$ & $100/100$ & $100/100$ & $99/100$ & $97/100$ \\
9 & $97/100$ & $95/100$ & $94/100$ & $94/100$ & $99/100$ & $99/100$ & $98/100$ & $100/100$ \\
\midrule
Suite SR (\%) & 95.50 & 95.20 & 96.90 & 97.20 & 99.50 & 99.60 & 97.10 & 97.30 \\
$\Delta$ (pp) & \multicolumn{2}{c}{$-0.30$} & \multicolumn{2}{c}{$+0.30$}
& \multicolumn{2}{c}{$+0.10$} & \multicolumn{2}{c}{$+0.20$} \\
\bottomrule
\end{tabular}
\end{table}

\begin{table}[!htbp]
\centering
\small
\setlength{\tabcolsep}{5pt}
\renewcommand{\arraystretch}{1.08}
\caption{\textbf{Action averaging for Cosmos3-Nano on LIBERO-10.}
Entries give successes / completed rollouts. Task IDs follow the
LIBERO-10 ordering in Table~\ref{tab:app-libero-goal-long-tasks}.}
\label{tab:flow-averaging-cosmos}
\begin{tabularx}{\linewidth}{@{}rXrr@{}}
\toprule
ID & Task & $K=1$ & $K=16$ mean \\
\midrule

0 & Put both the alphabet soup and the tomato sauce in the basket. & $8/10$ & $9/10$ \\
1 & Put both the cream cheese box and the butter in the basket. & $10/10$ & $10/10$ \\
2 & Turn on the stove and put the moka pot on it. & $9/10$ & $10/10$ \\
3 & Put the black bowl in the bottom drawer of the cabinet and close it. & $10/10$ & $10/10$ \\
4 & Put the white mug on the left plate and the yellow and white mug on the right plate. & $10/10$ & $10/10$ \\
5 & Pick up the book and place it in the back compartment of the caddy. & $10/10$ & $10/10$ \\
6 & Put the white mug on the plate and the chocolate pudding to the right of the plate. & $10/10$ & $9/10$ \\
7 & Put both the alphabet soup and the cream cheese box in the basket. & $10/10$ & $10/10$ \\
8 & Put both moka pots on the stove. & $9/10$ & $10/10$ \\
9 & Put the yellow and white mug in the microwave and close it. & $10/10$ & $10/10$ \\
\midrule
\multicolumn{2}{@{}l}{Mean success (\%)} & 96.00 & 98.00 \\
\bottomrule
\end{tabularx}
\end{table}

\FloatBarrier

\Needspace{12\baselineskip}
\subsection{Exploration with Controllable Sampling Noise}
\label{sec:appendix-exploration}

We evaluate stochastic exploration on state-based Tool-Hang with
Chi-Transformer. We compare native Flow-Policy sampling with Gaussian
perturbations of HT-Policy action predictions, scaled either by the learned
$\sigma(o)$ or by a fixed scale matched to the noiseless baseline.
We measure exploration using pass@$k$, the estimated probability of
at least one successful rollout within $k$ attempts.

\begin{table}[!htbp]
\centering
\caption{\textbf{Exploration pass@$k$ (\%) on state-based Tool-Hang with Chi-Transformer.}
Estimates are computed per initial state from eight rollouts and averaged
over 100 states. $\alpha$ controls HT-Policy sampling noise.
Dashes indicate unreported values or an inapplicable noise multiplier.}
\label{tab:exploration-passk}
\small
\setlength{\tabcolsep}{5pt}
\renewcommand{\arraystretch}{1.1}
\begin{tabular*}{\linewidth}{@{\extracolsep{\fill}}lcrrrr@{}}
\toprule
Method & $\alpha$ & pass@1 & pass@2 & pass@4 & pass@8 \\
\midrule
Flow-Policy (native) & --- & 76.25 & 93.68 & 99.67 & 100.00 \\
HT-Policy (no noise) & 0 & 81.00 & --- & --- & --- \\
\midrule
HT-Policy ($\sigma$-scaled) & 0.125 & 79.38 & 95.61 & 99.76 & 100.00 \\
HT-Policy (fixed scale) & 0.125 & 78.88 & 95.39 & 99.81 & 100.00 \\
\addlinespace[2pt]
HT-Policy ($\sigma$-scaled) & 0.25 & 78.63 & 95.82 & 99.90 & 100.00 \\
HT-Policy (fixed scale) & 0.25 & 76.88 & 94.25 & 99.56 & 100.00 \\
\addlinespace[2pt]
HT-Policy ($\sigma$-scaled) & 0.5 & 78.13 & 94.68 & 99.64 & 100.00 \\
HT-Policy (fixed scale) & 0.5 & 79.38 & 95.79 & 99.93 & 100.00 \\
\addlinespace[2pt]
HT-Policy ($\sigma$-scaled) & 1 & 75.88 & 94.43 & 99.76 & 100.00 \\
HT-Policy (fixed scale) & 1 & 78.75 & 95.11 & 99.60 & 100.00 \\
\addlinespace[2pt]
HT-Policy ($\sigma$-scaled) & 2 & 67.13 & 89.46 & 99.16 & 100.00 \\
HT-Policy (fixed scale) & 2 & 54.88 & 79.18 & 95.71 & 100.00 \\
\bottomrule
\end{tabular*}
\end{table}

Table~\ref{tab:exploration-passk} shows that modest noise gives
HT-Policies pass@$k$ comparable to Flow-Policies.
Both stochastic HT-Policy variants reach 100\% at pass@8 across the tested
noise strengths, showing that
direct action prediction can support successful stochastic exploration
in this setting.
\FloatBarrier

\subsection{Policy Training and Evaluation Settings}
\label{sec:appendix-training}

This section specifies the data, trainable modules, and evaluation
protocols for Sections~\ref{sec:evaluation-scratch}
and~\ref{sec:evaluaiton-generality}.
Throughout this appendix, $H$ is the supervised action-chunk length,
$D_a$ is the number of valid action channels, and $d=HD_a$.

For Table~\ref{tab:rq2-performance}, we evaluate each trained model
using ten evaluation seeds and report the
mean success rate and standard error across seeds.
Table~\ref{tab:app-evaluation} lists the
rollout budgets per task and seed, executed action horizons, and
Flow-Policy sampling steps. MSE-Policies and HT-Policies predict actions
in one forward pass.

\subsubsection{Real-Robot Experiments}
\label{sec:appendix-real-robot}

Push-T, Peel-Note, and Insert-T use a PiperX robot with a
front-view Logitech C920s camera. Each task has 100 demonstrations
collected using Quest 3-assisted VR teleoperation.
All methods use the same demonstrations and train U-Net policies
from scratch. The $640\times480$ RGB images are cropped to
$480\times480$ and resized to $256\times256$.
Policies also receive a ten-dimensional state: end-effector
position, 6D orientation, and gripper opening width.

In Push-T, the robot pushes a T-shaped object into alignment with
a target pose. The object's initial position is sampled within a
$10\times10$~cm bounding box, with an in-plane rotation sampled
from $[-30^\circ,30^\circ]$.

In Insert-T, the robot grasps and lifts a T-shaped object, then
places it into a holder. The object's initial position is sampled within a
$16\times5$~cm region to the right of the holder, with an in-plane
rotation sampled from $[-45^\circ,45^\circ]$.

In Peel-Note, the robot uses its gripper to peel a note off the
tabletop. The note's initial position is sampled within a
$5\times5$~cm region.

The control loop executes actions at 30 Hz. Each policy call
predicts a 16-step chunk, of which the first eight actions are
executed before replanning.
Push-T uses the gripper instead of DP's cylindrical end effector.
For a fair comparison, all methods use the same gripper during
training and evaluation.
Each method is evaluated in 50 rollouts per task.

\subsubsection{Scale Prediction}
\label{sec:appendix-scale-head}

Each HT-Policy predicts one positive scalar scale $\sigma_\theta(o)$ for each
observation and action chunk, shared across its $d$ valid action
coordinates. The scale head is trained jointly with the action
predictor through the HT loss. At inference, only the predicted
actions are used.

\paragraph{\texorpdfstring{$\pi_{0.5}$}{pi0.5}.}
The scale head first pools its raw outputs into a scalar and then
applies softplus. Its final-layer bias is initialized to $-0.396$.

\paragraph{GR00T N1.7.}
The scale head applies softplus before pooling. Let $z_j(o)$ be its
raw output for action coordinate $j$, and let $m_j$ indicate whether
that coordinate is supervised. The scalar scale is
\[
    \sigma_\theta(o)
    =\frac{\sum_j m_j\operatorname{softplus}(z_j(o))}{\sum_j m_j}
    +10^{-3}.
\]
The average uses only supervised coordinates. We initialize the
scale to the action-residual RMS before fine-tuning, with biases
$-1.088$, $-0.4535$, and $-0.5093$ for RoboCasa-GR1, Bridge, and
Fractal, respectively.

\paragraph{Cosmos3-Nano.}
The scale head maps action features to a positive scalar scale
through softplus. Its final-layer weights are initialized to zero.
We calibrate the initial bias on 64 examples to match the
residual RMS, giving $b_0=-0.17432$.

\paragraph{Learning rate.}
The scale head shares the network's learning rate in all settings
except Fractal. There, it uses $1.5\times10^{-4}$,
while all other parameter groups use $10^{-4}$.

\subsubsection{\texorpdfstring{$\pi_{0.5}$}{pi0.5} on LIBERO}

\begin{table}[htbp]
\centering
\caption{\textbf{Training budgets for the large-model experiments.} Batch size
is global. Cosmos3-Nano batch size depends on packing.}
\label{tab:app-training}
\small
\setlength{\tabcolsep}{5pt}
\renewcommand{\arraystretch}{1.1}
\begin{tabularx}{\linewidth}{@{}Xrrrrr@{}}
\toprule
Model / dataset & Updates & Batch & Base LR & Warmup & $\nu$ \\
\midrule
$\pi_{0.5}$ / LIBERO & 30,000 & 32 & $2.5\!\times\!10^{-5}$ & 1,000 & 1,400 \\
GR00T / RoboCasa-GR1 & 60,000 & 512 & $10^{-4}$ & 3,000 & 928 \\
GR00T / Bridge & 20,000 & 1,024 & $10^{-4}$ & 1,000 & 224 \\
GR00T / Fractal & 20,000 & 1,024 & $10^{-4}$ & 1,000 & 224 \\
Cosmos3-Nano / LIBERO-10 & 2,000 & $\leq2,048$ & $5\!\times\!10^{-5}$ & 500 & 640 \\
\bottomrule
\end{tabularx}
\end{table}

\begin{table}[htbp]
\centering
\caption{\textbf{Training data and effective action dimensions.} Episode and
frame counts refer to the datasets used for training.}
\label{tab:app-data-actions}
\small
\setlength{\tabcolsep}{6pt}
\renewcommand{\arraystretch}{1.1}
\begin{tabularx}{\linewidth}{@{}Xrrrrr@{}}
\toprule
Dataset & Episodes & Frames & $H$ & $D_a$ & $d$ \\
\midrule
LIBERO, four suites ($\pi_{0.5}$) & 1,693 & 273,465 & 50 & 7 & 350 \\
RoboCasa-GR1 & 24,000 & 5,820,277 & 8 & 29 & 232 \\
Bridge & 53,192 & 1,893,026 & 8 & 7 & 56 \\
Fractal & 87,212 & 3,786,400 & 8 & 7 & 56 \\
LIBERO-10 (Cosmos3-Nano) & 379 & 101,469 & 16 & 10 & 160 \\
\bottomrule
\end{tabularx}
\end{table}

\paragraph{Training.}
We train MSE-Policies and HT-Policies using the LeRobot fine-tuning recipe.
Fine-tuning updates the full vision, language, and action-expert network
with global batch size 32, BF16 precision, and no EMA.
AdamW uses $(\beta_1,\beta_2)=(0.9,0.95)$, $\epsilon=10^{-8}$,
weight decay 0.01, and gradient clipping at norm 1.
After 1,000 warmup steps, the learning rate decays from
$2.5\times10^{-5}$ to $2.5\times10^{-6}$ over the 30,000-step run.

\paragraph{Inputs and action prediction.}
Training covers all 40 LIBERO tasks (Table~\ref{tab:app-data-actions}).
Inputs contain two RGB views and an eight-dimensional state.
Images are resized and padded to
$224\times224$; this recipe disables training image augmentation.
State and action channels use quantile normalization.
The action chunk contains 50 seven-dimensional actions, so
$d=350$ and the main HT loss setting is $\nu=1400$.
HT-Policies and MSE-Policies use zero action input and timestep zero to
predict clean actions in one forward pass. The Flow-Policy uses ten Euler steps.

\begin{table}[htbp]
\centering
\caption{\textbf{Evaluation settings for the 10-seed comparison
in Table~\ref{tab:rq2-performance}.}
Rollout budgets are per task and evaluation seed;
actual GR00T rollout counts can slightly exceed these budgets.
``Execute'' is the number of actions applied before replanning.
MSE-Policies and HT-Policies each use a single forward pass.}
\label{tab:app-evaluation}
\small
\setlength{\tabcolsep}{4pt}
\renewcommand{\arraystretch}{1.1}
\begin{tabularx}{\linewidth}{@{}Xrrrrr@{}}
\toprule
Model / benchmark & Tasks & \makecell[r]{Rollout budget/\\task/seed}
& Execute & \makecell[r]{Flow\\steps} & \makecell[r]{Step\\limit} \\
\midrule
$\pi_{0.5}$ / Spatial and Object & $2\times10$ & 100 & 10 & 10 & 280 \\
$\pi_{0.5}$ / Goal & 10 & 100 & 10 & 10 & 300 \\
$\pi_{0.5}$ / LIBERO-10 & 10 & 100 & 10 & 10 & 520 \\
GR00T / RoboCasa-GR1 & 24 & 20 & 8 & 4 & 720 \\
GR00T / Bridge--WidowX & 7 & 50 & 4 & 4 & 300 \\
GR00T / Fractal--Google Robot & 6 & 100 & 1 & 4 & 300 \\
Cosmos3-Nano / LIBERO-10 & 10 & 10 & 16 & 30 & 520 \\
\bottomrule
\end{tabularx}
\end{table}

\paragraph{Evaluation.}
Evaluation covers LIBERO-Spatial, LIBERO-Object, LIBERO-Goal,
and LIBERO-10, with ten tasks in each suite.
The environment uses the official initial-state banks and ten
no-op settling steps. Each task is evaluated in
100 rollouts per evaluation seed.
The policy executes ten actions before replanning.
Episode limits are 280 steps for Spatial and Object, 300 for
Goal, and 520 for LIBERO-10.
The reported four-suite score averages the task rates within
each suite and then averages the four suites, giving equal
weight to all 40 tasks.

\subsubsection{GR00T N1.7 on RoboCasa-GR1 and SIMPLER}
\label{sec:appendix-gr1-training}

\paragraph{Trainable parameters.}
GR00T N1.7 uses a Cosmos-Reason2-2B vision-language backbone.
The pretrained vision-language backbone is frozen. Training updates the
action transformer, state and action projections, action decoder,
and associated normalization layers.

\paragraph{Optimization.}
Training uses AdamW, cosine learning-rate
decay, weight decay $10^{-5}$, and gradient clipping at norm 1.
We use BF16 precision.
Table~\ref{tab:app-training} gives the batch sizes, budgets, and
warmup lengths. The state dropout probabilities are 0.2 on
RoboCasa-GR1, 0.8 on Bridge, and 0.5 on Fractal.
The HT loss uses $\nu=4d$.

\paragraph{RoboCasa-GR1 data and actions.}
Training uses 1,000 demonstrations for each of 24 tasks,
for a total of 24,000 episodes at 20 Hz.
Each observation contains the current ego-view image, robot state,
and task text. Each action has 29 supervised channels:
seven for each arm, six for each hand, and three for the
waist. Arm and hand actions are expressed relative to the current
state; waist actions remain absolute.
The supervised action chunk has eight steps, giving $d=8\times29=232$.

\paragraph{SIMPLER data and actions.}
BridgeData V2 and Fractal/RT-1 provide the demonstrations for
WidowX and Google Robot, respectively (Table~\ref{tab:app-data-actions}).
Both policies receive one RGB view, robot state, and instruction text.
Both predict eight actions with seven channels each: three
translation components, three rotation components, and gripper
control. Their effective dimension is $d=56$.
Both datasets retain their incremental-action conventions during
training and execution.

\paragraph{Preprocessing.}
GR00T uses percentile normalization, outlier clipping, and statistics
computed from the fine-tuning data. Images are resized to a
shortest edge of 256 pixels and cropped to 95\% of that size.
Training uses random crops and color jitter; evaluation uses
center crops and the VLM's image processor.
The full training datasets are used, with no separate policy
validation dataset.

\paragraph{Closed-loop evaluation.}
On RoboCasa-GR1, each evaluation seed targets 20 rollouts per task, allows up to
720 environment steps, and executes eight actions before replanning.

SIMPLER uses the standard benchmark environments. The rollout budgets per task and seed
are 50 for the seven Bridge tasks and 100 for the six Fractal tasks.
Both use a 300-step limit, but WidowX executes four predicted
actions before replanning and Google Robot executes one.

Actual rollout counts can slightly exceed the requested budgets.
Each task's success rate uses the number of completed rollouts as
its denominator. We average task rates equally within each seed,
then report the mean and standard error across the ten evaluation seeds
in Table~\ref{tab:rq2-performance}.

WidowX images are resized to $256\times256$, and Google Robot
images to $256\times320$, before policy preprocessing.
We follow the benchmark action and gripper conventions.
HT-Policies and MSE-Policies execute their predicted means;
Flow-Policies use four sampling steps.

\subsubsection{From-Scratch Policies on RoboCasa-GR1}
\label{sec:appendix-gr1-scratch}

The from-scratch experiments in Section~\ref{sec:evaluation-scratch}
use the same GR00T N1.7 architecture, data, optimization, and
evaluation protocols described in Appendix~\ref{sec:appendix-gr1-training},
with a randomly initialized action head. The pretrained vision-language
backbone remains frozen. Thus, ``from scratch'' refers to learning
the action predictor from fixed VLM features, as in the main text.

\subsubsection{Cosmos3-Nano on LIBERO-10}
\label{sec:appendix-cosmos-training}

\paragraph{Joint video and action training.}
We retain the video flow-matching objective and train the action
branch with the HT or MSE loss. The action branch receives zero
action input at timestep zero and predicts clean normalized actions.
Both branch losses have weight 10, with the standard auxiliary terms
unchanged. Training updates the shared generation expert and modality
projections, so video supervision also affects action prediction.

\paragraph{Optimization.}
Training uses FusedAdam with BF16 precision, base learning rate
$5\times10^{-5}$, and an action learning-rate multiplier of five.
The schedule has 500 warmup steps and a 16,000-step cycle.
We train for 2,000 updates with effective batch size up to 2,048
(Table~\ref{tab:app-training}).

\paragraph{Actions and evaluation.}
The policy uses agent-view and wrist images from LIBERO-10.
Actions have ten valid channels: translation, a six-dimensional
rotation representation, and gripper control. A 16-step chunk
gives $d=160$ and $\nu=640$.
At inference, one joint forward pass uses the observed inputs,
zero action input, and video noise to predict actions.
No future ground-truth video or video generation is required.
Video noise can affect actions through joint attention, so the
policy remains stochastic despite using a single action-prediction pass.

The protocol uses the official LIBERO-10 initial states, ten
settling steps and ten rollouts
per task and evaluation seed. All 16 predicted actions are executed
before replanning, with a 520-step episode limit.
The Flow-Policy uses its native 30-step sampler.

\FloatBarrier

\subsection{Additional Comparisons with Generative Policies}
\label{sec:appendix-comparisons}

\subsubsection{Comparison with Diffusion Policy}
\label{sec:appendix-dp}

Tables~\ref{tab:appendix-dp-state} and~\ref{tab:appendix-dp-image}
give the task-level results for DP and HT-Policies with state and image observations
used in Table~\ref{tab:rq1-performance}.
DP-C and DP-T are the U-Net and Transformer variants
reported by~\citet{chi2023diffusionpolicy}; HT-Policies use Chi-UNet
and Chi-Transformer, respectively.
The HT-Policy runs use three training seeds and 22 rollouts per evaluation,
with fixed initial conditions. DP also used 22 initial states for
RoboMimic, as documented in the evaluation-code footnote in
Section 5.2 of its extended paper.

Each cell reports best/last10: the highest evaluation score during
training and the mean of the final ten evaluations, averaged over
three training seeds.
The RoboMimic mean covers its nine tasks; Push-T is listed separately
and reports normalized coverage.
Table~\ref{tab:rq1-performance} averages the state and image results
equally for each backbone. Figure~\ref{fig:teaser} further averages
the best scores across the two backbones.
All averages are computed before rounding.

\begin{table}[!htbp]
\centering
\caption{\textbf{DP and HT with state observations.}
Bold marks the highest value per row and metric across both backbones, before rounding.}
\label{tab:appendix-dp-state}
\small
\setlength{\tabcolsep}{7pt}
\renewcommand{\arraystretch}{1.08}
\begin{tabular*}{\linewidth}{@{\extracolsep{\fill}}lcccc@{}}
\toprule
& \multicolumn{2}{c}{U-Net} & \multicolumn{2}{c}{Transformer} \\
\cmidrule(lr){2-3}\cmidrule(lr){4-5}
Task & DP-C & HT & DP-T & HT \\
& best/last10 & best/last10 & best/last10 & best/last10 \\
\midrule
Lift-PH & \textbf{1.00}/0.98 & \textbf{1.00}/\textbf{1.00} & \textbf{1.00}/\textbf{1.00} & \textbf{1.00}/0.93 \\
Lift-MH & \textbf{1.00}/0.97 & \textbf{1.00}/\textbf{1.00} & \textbf{1.00}/\textbf{1.00} & \textbf{1.00}/0.96 \\
Can-PH & \textbf{1.00}/0.96 & \textbf{1.00}/\textbf{1.00} & \textbf{1.00}/\textbf{1.00} & \textbf{1.00}/0.94 \\
Can-MH & \textbf{1.00}/0.96 & \textbf{1.00}/\textbf{0.98} & \textbf{1.00}/0.94 & \textbf{1.00}/\textbf{0.98} \\
Square-PH & \textbf{1.00}/0.93 & \textbf{1.00}/\textbf{0.97} & \textbf{1.00}/0.89 & 0.97/0.88 \\
Square-MH & 0.97/0.82 & \textbf{0.98}/\textbf{0.88} & 0.95/0.81 & 0.88/0.74 \\
Transport-PH & 0.94/0.82 & 0.91/0.78 & \textbf{1.00}/\textbf{0.84} & 0.70/0.48 \\
Transport-MH & 0.68/\textbf{0.46} & \textbf{0.79}/0.45 & 0.62/0.35 & 0.36/0.14 \\
Tool-Hang & 0.50/0.30 & 0.88/0.65 & \textbf{1.00}/\textbf{0.87} & 0.97/0.80 \\
\midrule
RoboMimic mean & 0.899/0.800 & 0.951/\textbf{0.857} & \textbf{0.952}/0.856 & 0.876/0.761 \\
\midrule
Push-T & 0.95/0.91 & 0.96/0.93 & 0.95/0.79 & \textbf{1.00}/\textbf{0.98} \\
\bottomrule
\end{tabular*}
\end{table}

\begin{table}[!htbp]
\centering
\caption{\textbf{DP and HT with image observations.}
Bold marks the highest value per row and metric across both backbones, before rounding.}
\label{tab:appendix-dp-image}
\small
\setlength{\tabcolsep}{7pt}
\renewcommand{\arraystretch}{1.08}
\begin{tabular*}{\linewidth}{@{\extracolsep{\fill}}lcccc@{}}
\toprule
& \multicolumn{2}{c}{U-Net} & \multicolumn{2}{c}{Transformer} \\
\cmidrule(lr){2-3}\cmidrule(lr){4-5}
Task & DP-C & HT & DP-T & HT \\
& best/last10 & best/last10 & best/last10 & best/last10 \\
\midrule
Lift-PH & \textbf{1.00}/\textbf{1.00} & \textbf{1.00}/0.95 & \textbf{1.00}/\textbf{1.00} & \textbf{1.00}/\textbf{1.00} \\
Lift-MH & \textbf{1.00}/\textbf{1.00} & \textbf{1.00}/0.95 & \textbf{1.00}/0.99 & \textbf{1.00}/0.99 \\
Can-PH & \textbf{1.00}/0.97 & \textbf{1.00}/0.94 & \textbf{1.00}/0.98 & \textbf{1.00}/\textbf{1.00} \\
Can-MH & \textbf{1.00}/0.96 & \textbf{1.00}/0.92 & \textbf{1.00}/\textbf{0.98} & \textbf{1.00}/0.95 \\
Square-PH & 0.98/0.92 & \textbf{1.00}/0.95 & \textbf{1.00}/0.90 & \textbf{1.00}/\textbf{0.96} \\
Square-MH & 0.98/0.84 & \textbf{1.00}/\textbf{0.85} & 0.94/0.80 & 0.95/\textbf{0.85} \\
Transport-PH & \textbf{1.00}/\textbf{0.93} & \textbf{1.00}/0.92 & 0.98/0.81 & 0.95/0.90 \\
Transport-MH & \textbf{0.89}/\textbf{0.69} & 0.82/0.56 & 0.73/0.50 & 0.86/0.63 \\
Tool-Hang & \textbf{0.95}/0.73 & 0.77/0.65 & 0.76/0.47 & 0.91/\textbf{0.79} \\
\midrule
RoboMimic mean & \textbf{0.978}/0.893 & 0.954/0.854 & 0.934/0.826 & 0.963/\textbf{0.897} \\
\midrule
Push-T & 0.91/0.84 & 0.93/0.89 & 0.78/0.66 & \textbf{0.95}/\textbf{0.91} \\
\bottomrule
\end{tabular*}
\end{table}

\FloatBarrier

\subsubsection{Comparison with MIP}
\label{sec:appendix-mip}

\paragraph{Comparison setup.}
We compare HT-Policies with Flow-Policies, MSE-Policies, Straight Flow,
and MIP on nine RoboMimic tasks and Push-T.
All baseline scores are taken from Tables 11 and 12 of~\citet{pan2026much}.
Our tables label that work's Regression baseline as MSE.
The RoboMimic labels PH and MH denote proficient-human and
multi-human demonstrations.
Each method is evaluated with three backbones: Sudeep-DiT,
Chi-Transformer, and Chi-UNet.
Tables~\ref{tab:appendix-mip-state} and~\ref{tab:appendix-mip-comparison}
report results for state and image observations.
Each cell gives best/last5: the highest evaluation score during
training and the mean of the final five evaluations, averaged over
three training seeds.
The Mean column reports the average score over the ten tasks.
RoboMimic uses binary success, while Push-T uses normalized coverage.

\paragraph{Performance comparison.}
HT-Policies use a single-pass regression head and perform on par with MIP overall.
The bottom block of each table reports the best score across
backbones for each method, task, and metric.
Averaged over the ten tasks and both observation types, these scores
are $91.3\%/85.5\%$ for HT-Policies and $91.5\%/86.1\%$ for MIP.
HT-Policies score slightly lower on state observations and slightly higher
on image observations.

\begin{table}[!htbp]
\centering
\fontsize{7.5}{9}\selectfont
\setlength{\tabcolsep}{1.5pt}
\renewcommand{\arraystretch}{1.04}
\setlength{\aboverulesep}{2pt}
\setlength{\belowrulesep}{2pt}
\caption{\textbf{Scores on state-based tasks.} Entries report best/last5, averaged over three seeds.
RoboMimic reports success; Push-T reports normalized coverage.
Baselines follow Table 11 of~\citet{pan2026much}.
Bold marks the best value per task and metric across all methods and backbones, before rounding.}
\label{tab:appendix-mip-state}
\begin{tabular}{@{}l*{11}{c}@{}}
\toprule[0.6pt]
Method & \multicolumn{2}{c}{Lift} & \multicolumn{2}{c}{Can}
& \multicolumn{2}{c}{Square} & \multicolumn{2}{c}{Transport}
& Tool-Hang & Push-T & Mean \\
\midrule[0.3pt]
& PH & MH & PH & MH & PH & MH & PH & MH & & & \\
\midrule[0.3pt]
\multicolumn{12}{@{}l}{\textbf{Sudeep-DiT}} \\
Flow & \textbf{1.00}/\textbf{1.00} & \textbf{1.00}/0.99 & \textbf{1.00}/\textbf{1.00} & \textbf{1.00}/0.94 & \textbf{1.00}/\textbf{0.94} & 0.88/0.75 & 0.80/0.70 & 0.40/0.27 & 0.86/0.75 & \textbf{0.98}/0.95 & 0.89/0.83 \\
MSE & \textbf{1.00}/\textbf{1.00} & \textbf{1.00}/0.99 & \textbf{1.00}/0.98 & 0.92/0.90 & 0.94/0.86 & 0.72/0.53 & 0.50/0.44 & 0.12/0.06 & 0.52/0.39 & 0.92/0.83 & 0.76/0.70 \\
Straight Flow & \textbf{1.00}/\textbf{1.00} & \textbf{1.00}/0.98 & \textbf{1.00}/0.99 & 0.96/0.90 & 0.96/0.93 & 0.72/0.66 & 0.56/0.48 & 0.20/0.14 & 0.70/0.59 & 0.90/0.86 & 0.80/0.75 \\
MIP & \textbf{1.00}/\textbf{1.00} & \textbf{1.00}/0.99 & \textbf{1.00}/\textbf{1.00} & 0.98/0.95 & 0.98/\textbf{0.94} & 0.90/0.81 & 0.76/0.68 & 0.44/0.38 & \textbf{0.92}/\textbf{0.88} & 0.95/0.92 & 0.89/0.86 \\
\textbf{HT (ours)} & \textbf{1.00}/0.98 & \textbf{1.00}/1.00 & \textbf{1.00}/0.97 & 0.98/0.90 & 0.89/0.82 & 0.72/0.63 & 0.54/0.45 & 0.33/0.21 & 0.65/0.56 & 0.97/\textbf{0.95} & 0.81/0.75 \\
\addlinespace[3pt]
\multicolumn{12}{@{}l}{\textbf{Chi-Transformer}} \\
Flow & \textbf{1.00}/\textbf{1.00} & \textbf{1.00}/\textbf{1.00} & \textbf{1.00}/0.98 & \textbf{1.00}/0.93 & 0.96/0.89 & 0.78/0.74 & \textbf{0.88}/0.64 & 0.44/0.34 & 0.68/0.54 & 0.91/0.89 & 0.87/0.80 \\
MSE & \textbf{1.00}/0.99 & \textbf{1.00}/0.99 & \textbf{1.00}/0.96 & 0.98/0.92 & 0.92/0.85 & 0.74/0.61 & 0.68/0.51 & 0.28/0.20 & 0.40/0.36 & 0.93/0.88 & 0.79/0.73 \\
Straight Flow & \textbf{1.00}/\textbf{1.00} & \textbf{1.00}/0.99 & \textbf{1.00}/0.99 & 0.98/0.92 & 0.96/0.89 & 0.68/0.58 & 0.62/0.54 & 0.24/0.16 & 0.60/0.55 & 0.94/0.90 & 0.80/0.75 \\
MIP & \textbf{1.00}/\textbf{1.00} & \textbf{1.00}/\textbf{1.00} & \textbf{1.00}/\textbf{1.00} & 0.96/0.95 & 0.96/0.89 & 0.86/0.73 & 0.80/0.68 & 0.42/0.37 & 0.76/0.69 & 0.94/0.92 & 0.87/0.82 \\
\textbf{HT (ours)} & \textbf{1.00}/0.98 & \textbf{1.00}/0.99 & \textbf{1.00}/0.96 & \textbf{1.00}/0.95 & 0.95/0.86 & 0.72/0.60 & 0.63/0.48 & 0.33/0.22 & 0.88/0.81 & 0.96/0.93 & 0.85/0.78 \\
\addlinespace[3pt]
\multicolumn{12}{@{}l}{\textbf{Chi-UNet}} \\
Flow & \textbf{1.00}/\textbf{1.00} & \textbf{1.00}/\textbf{1.00} & \textbf{1.00}/\textbf{1.00} & \textbf{1.00}/\textbf{0.98} & 0.98/\textbf{0.94} & 0.90/0.78 & 0.80/\textbf{0.73} & 0.52/0.40 & 0.84/0.70 & \textbf{0.98}/0.94 & 0.90/0.85 \\
MSE & \textbf{1.00}/\textbf{1.00} & \textbf{1.00}/\textbf{1.00} & \textbf{1.00}/0.99 & \textbf{1.00}/0.96 & \textbf{1.00}/0.91 & \textbf{0.94}/\textbf{0.82} & 0.64/0.55 & 0.22/0.16 & 0.68/0.64 & 0.95/0.89 & 0.84/0.79 \\
Straight Flow & \textbf{1.00}/\textbf{1.00} & \textbf{1.00}/\textbf{1.00} & \textbf{1.00}/0.99 & \textbf{1.00}/0.92 & 0.98/0.90 & \textbf{0.94}/0.79 & 0.64/0.52 & 0.22/0.15 & 0.50/0.00 & 0.93/0.88 & 0.82/0.72 \\
MIP & \textbf{1.00}/\textbf{1.00} & \textbf{1.00}/\textbf{1.00} & \textbf{1.00}/0.99 & \textbf{1.00}/\textbf{0.98} & \textbf{1.00}/\textbf{0.94} & 0.92/0.81 & 0.80/0.69 & 0.62/0.46 & 0.80/0.64 & 0.97/0.95 & 0.91/0.85 \\
\textbf{HT (ours)} & \textbf{1.00}/1.00 & \textbf{1.00}/\textbf{1.00} & \textbf{1.00}/0.99 & \textbf{1.00}/0.97 & 0.98/0.92 & 0.87/0.74 & 0.78/0.67 & \textbf{0.63}/\textbf{0.49} & 0.72/0.56 & 0.93/0.91 & 0.89/0.82 \\
\midrule[0.3pt]
\multicolumn{12}{@{}l}{\textbf{Best backbone}} \\
Flow & \textbf{1.00}/\textbf{1.00} & \textbf{1.00}/\textbf{1.00} & \textbf{1.00}/\textbf{1.00} & \textbf{1.00}/\textbf{0.98} & \textbf{1.00}/\textbf{0.94} & 0.90/0.78 & \textbf{0.88}/\textbf{0.73} & 0.52/0.40 & 0.86/0.75 & \textbf{0.98}/0.95 & 0.91/0.85 \\
MSE & \textbf{1.00}/\textbf{1.00} & \textbf{1.00}/\textbf{1.00} & \textbf{1.00}/0.99 & \textbf{1.00}/0.96 & \textbf{1.00}/0.91 & \textbf{0.94}/\textbf{0.82} & 0.68/0.55 & 0.28/0.20 & 0.68/0.64 & 0.95/0.89 & 0.85/0.80 \\
Straight Flow & \textbf{1.00}/\textbf{1.00} & \textbf{1.00}/\textbf{1.00} & \textbf{1.00}/0.99 & \textbf{1.00}/0.92 & 0.98/0.93 & \textbf{0.94}/0.79 & 0.64/0.54 & 0.24/0.16 & 0.70/0.59 & 0.94/0.90 & 0.84/0.78 \\
MIP & \textbf{1.00}/\textbf{1.00} & \textbf{1.00}/\textbf{1.00} & \textbf{1.00}/\textbf{1.00} & \textbf{1.00}/\textbf{0.98} & \textbf{1.00}/\textbf{0.94} & 0.92/0.81 & 0.80/0.69 & 0.62/0.46 & \textbf{0.92}/\textbf{0.88} & 0.97/0.95 & 0.92/0.87 \\
\textbf{HT (ours)} & \textbf{1.00}/1.00 & \textbf{1.00}/\textbf{1.00} & \textbf{1.00}/0.99 & \textbf{1.00}/0.97 & 0.98/0.92 & 0.87/0.74 & 0.78/0.67 & \textbf{0.63}/\textbf{0.49} & 0.88/0.81 & 0.97/\textbf{0.95} & 0.91/0.85 \\
\bottomrule[0.6pt]
\end{tabular}
\end{table}

\begin{table}[!htbp]
\centering
\fontsize{7.5}{9}\selectfont
\setlength{\tabcolsep}{1.5pt}
\renewcommand{\arraystretch}{1.04}
\setlength{\aboverulesep}{2pt}
\setlength{\belowrulesep}{2pt}
\caption{\textbf{Scores on image-based tasks.} Entries report best/last5, averaged over three seeds.
RoboMimic reports success; Push-T reports normalized coverage.
Baselines follow Table 12 of~\citet{pan2026much}.
Bold marks the best value per task and metric across all methods and backbones, before rounding.}
\label{tab:appendix-mip-comparison}
\begin{tabular}{@{}l*{11}{c}@{}}
\toprule[0.6pt]
Method & \multicolumn{2}{c}{Lift} & \multicolumn{2}{c}{Can}
& \multicolumn{2}{c}{Square} & \multicolumn{2}{c}{Transport}
& Tool-Hang & Push-T & Mean \\
\midrule[0.3pt]
& PH & MH & PH & MH & PH & MH & PH & MH & & & \\
\midrule[0.3pt]
\multicolumn{12}{@{}l}{\textbf{Sudeep-DiT}} \\
Flow & \textbf{1.00}/\textbf{1.00} & \textbf{1.00}/\textbf{1.00} & \textbf{1.00}/0.99 & 0.96/0.94 & 0.96/\textbf{0.94} & 0.82/0.76 & 0.84/0.83 & 0.32/0.20 & \textbf{0.78}/0.57 & \textbf{0.92}/\textbf{0.89} & 0.86/0.81 \\
MSE & \textbf{1.00}/\textbf{1.00} & \textbf{1.00}/0.99 & \textbf{1.00}/\textbf{1.00} & 0.92/0.81 & 0.94/0.84 & 0.74/0.67 & 0.74/0.56 & 0.14/0.08 & 0.28/0.18 & 0.83/0.77 & 0.76/0.69 \\
Straight Flow & \textbf{1.00}/0.99 & \textbf{1.00}/0.99 & \textbf{1.00}/0.98 & 0.98/0.95 & \textbf{1.00}/0.93 & 0.82/0.72 & 0.86/0.83 & 0.26/0.19 & 0.46/0.40 & 0.85/0.79 & 0.82/0.78 \\
MIP & \textbf{1.00}/\textbf{1.00} & \textbf{1.00}/0.99 & \textbf{1.00}/0.98 & \textbf{1.00}/0.96 & \textbf{1.00}/0.92 & 0.90/0.83 & 0.90/0.84 & 0.50/0.31 & 0.76/\textbf{0.66} & 0.91/0.87 & 0.90/0.84 \\
\textbf{HT (ours)} & \textbf{1.00}/\textbf{1.00} & \textbf{1.00}/0.99 & \textbf{1.00}/0.98 & 0.99/0.94 & 0.96/0.88 & 0.83/0.73 & 0.90/0.83 & 0.72/0.58 & 0.75/0.66 & 0.91/0.87 & 0.91/0.85 \\
\addlinespace[3pt]
\multicolumn{12}{@{}l}{\textbf{Chi-Transformer}} \\
Flow & \textbf{1.00}/\textbf{1.00} & \textbf{1.00}/0.99 & \textbf{1.00}/0.96 & 0.98/0.92 & 0.98/0.93 & 0.70/0.66 & 0.80/0.77 & 0.24/0.22 & 0.54/0.40 & 0.89/0.85 & 0.81/0.77 \\
MSE & \textbf{1.00}/0.98 & \textbf{1.00}/0.98 & \textbf{1.00}/0.96 & \textbf{1.00}/0.94 & 0.98/0.90 & 0.76/0.70 & 0.94/0.87 & 0.40/0.27 & 0.44/0.36 & 0.85/0.81 & 0.84/0.78 \\
Straight Flow & \textbf{1.00}/\textbf{1.00} & \textbf{1.00}/\textbf{1.00} & \textbf{1.00}/0.98 & \textbf{1.00}/0.95 & 0.98/\textbf{0.94} & 0.90/0.78 & 0.86/0.70 & 0.32/0.25 & 0.36/0.28 & 0.86/0.80 & 0.83/0.77 \\
MIP & \textbf{1.00}/\textbf{1.00} & \textbf{1.00}/0.98 & \textbf{1.00}/0.98 & 0.96/0.91 & 0.90/0.04 & 0.72/0.21 & 0.86/0.69 & 0.18/0.06 & 0.60/0.48 & 0.87/0.83 & 0.81/0.62 \\
\textbf{HT (ours)} & \textbf{1.00}/0.99 & \textbf{1.00}/0.99 & \textbf{1.00}/0.97 & 0.99/0.93 & 0.98/0.90 & 0.85/0.75 & 0.90/0.83 & \textbf{0.78}/\textbf{0.66} & 0.54/0.38 & 0.89/0.82 & 0.89/0.82 \\
\addlinespace[3pt]
\multicolumn{12}{@{}l}{\textbf{Chi-UNet}} \\
Flow & \textbf{1.00}/\textbf{1.00} & \textbf{1.00}/\textbf{1.00} & \textbf{1.00}/0.98 & \textbf{1.00}/\textbf{0.97} & 0.96/0.90 & 0.90/0.79 & 0.78/0.61 & 0.24/0.16 & 0.48/0.37 & \textbf{0.92}/0.87 & 0.83/0.77 \\
MSE & \textbf{1.00}/0.99 & \textbf{1.00}/0.96 & 0.98/0.87 & 0.84/0.70 & 0.94/0.86 & 0.74/0.66 & 0.66/0.64 & 0.18/0.10 & 0.30/0.23 & 0.90/0.85 & 0.75/0.69 \\
Straight Flow & \textbf{1.00}/0.99 & \textbf{1.00}/0.94 & \textbf{1.00}/0.96 & 0.98/0.93 & 0.92/0.62 & 0.72/0.68 & 0.50/0.22 & 0.00/0.00 & 0.06/0.02 & 0.87/0.82 & 0.71/0.62 \\
MIP & \textbf{1.00}/\textbf{1.00} & \textbf{1.00}/\textbf{1.00} & \textbf{1.00}/0.98 & \textbf{1.00}/0.95 & 0.96/0.91 & \textbf{0.92}/\textbf{0.84} & \textbf{0.96}/\textbf{0.91} & 0.52/0.37 & 0.56/0.50 & 0.83/0.78 & 0.88/0.82 \\
\textbf{HT (ours)} & \textbf{1.00}/0.97 & \textbf{1.00}/0.94 & 0.97/0.88 & 0.93/0.85 & 0.90/0.81 & 0.76/0.64 & 0.87/0.76 & 0.63/0.49 & 0.62/0.47 & 0.85/0.80 & 0.85/0.76 \\
\midrule[0.3pt]
\multicolumn{12}{@{}l}{\textbf{Best backbone}} \\
Flow & \textbf{1.00}/\textbf{1.00} & \textbf{1.00}/\textbf{1.00} & \textbf{1.00}/0.99 & \textbf{1.00}/\textbf{0.97} & 0.98/\textbf{0.94} & 0.90/0.79 & 0.84/0.83 & 0.32/0.22 & \textbf{0.78}/0.57 & \textbf{0.92}/\textbf{0.89} & 0.87/0.82 \\
MSE & \textbf{1.00}/\textbf{1.00} & \textbf{1.00}/0.99 & \textbf{1.00}/\textbf{1.00} & \textbf{1.00}/0.94 & 0.98/0.90 & 0.76/0.70 & 0.94/0.87 & 0.40/0.27 & 0.44/0.36 & 0.90/0.85 & 0.84/0.79 \\
Straight Flow & \textbf{1.00}/\textbf{1.00} & \textbf{1.00}/\textbf{1.00} & \textbf{1.00}/0.98 & \textbf{1.00}/0.95 & \textbf{1.00}/\textbf{0.94} & 0.90/0.78 & 0.86/0.83 & 0.32/0.25 & 0.46/0.40 & 0.87/0.82 & 0.84/0.80 \\
MIP & \textbf{1.00}/\textbf{1.00} & \textbf{1.00}/\textbf{1.00} & \textbf{1.00}/0.98 & \textbf{1.00}/0.96 & \textbf{1.00}/0.92 & \textbf{0.92}/\textbf{0.84} & \textbf{0.96}/\textbf{0.91} & 0.52/0.37 & 0.76/\textbf{0.66} & 0.91/0.87 & 0.91/0.85 \\
\textbf{HT (ours)} & \textbf{1.00}/\textbf{1.00} & \textbf{1.00}/0.99 & \textbf{1.00}/0.98 & 0.99/0.94 & 0.98/0.90 & 0.85/0.75 & 0.90/0.83 & \textbf{0.78}/\textbf{0.66} & 0.75/0.66 & 0.91/0.87 & 0.92/0.86 \\
\bottomrule[0.6pt]
\end{tabular}
\end{table}

\FloatBarrier

\subsubsection{Results for \texorpdfstring{$\pi_{0.5}$}{pi0.5} on LIBERO}
\label{sec:appendix-libero-results}

Table~\ref{tab:appendix-libero-suites} compares the official OpenPI
Flow-Policy scores with our Flow-Policy and HT-Policy results across the four
LIBERO suites. The policies use the LeRobot training recipe.
Tables~\ref{tab:app-libero-tasks} and~\ref{tab:app-libero-goal-long-tasks}
give the Flow-Policy and HT-Policy scores for all 40 tasks as mean $\pm$
standard error across 10 evaluation seeds for each trained model.
The overall means match Table~\ref{tab:rq2-performance}.

\begin{table}[!htbp]
\centering
\caption{\textbf{Success rates (\%) of $\pi_{0.5}$ across LIBERO suites.}
Flow and HT report mean $\pm$ standard error across 10 evaluation
seeds for each trained model.
Bold marks the higher mean of Flow and HT.}
\label{tab:appendix-libero-suites}
\small
\setlength{\tabcolsep}{6pt}
\renewcommand{\arraystretch}{1.1}
\begin{tabular*}{\linewidth}{@{\extracolsep{\fill}}lrrr@{}}
\toprule
Suite & Flow (OpenPI) & Flow & HT \\
\midrule
LIBERO-Spatial & 98.80 & $97.13\pm0.11$ & $\mathbf{97.71\pm0.13}$ \\
LIBERO-Object & 98.20 & $99.44\pm0.06$ & $\mathbf{99.80\pm0.00}$ \\
LIBERO-Goal & 98.00 & $96.55\pm0.19$ & $\mathbf{96.94\pm0.23}$ \\
LIBERO-10 & 92.40 & $\mathbf{95.23\pm0.14}$ & $93.44\pm0.06$ \\
\midrule
Mean & 96.85 & $\mathbf{97.09\pm0.04}$ & $96.97\pm0.07$ \\
\bottomrule
\end{tabular*}
\end{table}

\FloatBarrier
\begin{table}[!htbp]
\centering
\small
\setlength{\tabcolsep}{4pt}
\renewcommand{\arraystretch}{1.08}
\settowidth{\appendixtaskwidth}{ID$\mathbf{100.00\pm0.00}$$\mathbf{100.00\pm0.00}$}
\setlength{\appendixtaskwidth}{\dimexpr\linewidth-\appendixtaskwidth-6\tabcolsep\relax}
\caption{\textbf{Task-level success rates (\%) of $\pi_{0.5}$ on LIBERO-Spatial and LIBERO-Object.}
Task scores are mean $\pm$ standard error over 10 evaluation seeds for each trained model.
Bold marks the higher mean for each task.}
\label{tab:app-libero-tasks}
\begin{tabular}{@{}r>{\raggedright\arraybackslash}p{\appendixtaskwidth}rr@{}}
\toprule
ID & Task & Flow & HT \\
\midrule
\multicolumn{4}{@{}l}{\textbf{LIBERO-Spatial}} \\
0 & Pick up the black bowl between the plate and the ramekin and place it on the plate. & $\mathbf{99.90\pm0.10}$ & $\mathbf{99.90\pm0.10}$ \\
1 & Pick up the black bowl next to the ramekin and place it on the plate. & $\mathbf{99.60\pm0.22}$ & $99.30\pm0.21$ \\
2 & Pick up the black bowl from table center and place it on the plate. & $\mathbf{99.70\pm0.21}$ & $\mathbf{99.70\pm0.15}$ \\
3 & Pick up the black bowl on the cookie box and place it on the plate. & $\mathbf{99.40\pm0.22}$ & $98.80\pm0.36$ \\
4 & Pick up the black bowl in the top drawer of the wooden cabinet and place it on the plate. & $\mathbf{94.20\pm0.39}$ & $94.10\pm0.64$ \\
5 & Pick up the black bowl on the ramekin and place it on the plate. & $81.40\pm0.75$ & $\mathbf{87.70\pm0.97}$ \\
6 & Pick up the black bowl next to the cookie box and place it on the plate. & $\mathbf{100.00\pm0.00}$ & $99.80\pm0.13$ \\
7 & Pick up the black bowl on the stove and place it on the plate. & $99.30\pm0.26$ & $\mathbf{99.50\pm0.17}$ \\
8 & Pick up the black bowl next to the plate and place it on the plate. & $98.60\pm0.31$ & $\mathbf{99.00\pm0.33}$ \\
9 & Pick up the black bowl on the wooden cabinet and place it on the plate. & $99.20\pm0.25$ & $\mathbf{99.30\pm0.21}$ \\
\addlinespace[2pt]
\multicolumn{2}{@{}l}{Suite mean} & $97.13\pm0.11$ & $\mathbf{97.71\pm0.13}$ \\
\midrule
\multicolumn{4}{@{}l}{\textbf{LIBERO-Object}} \\
0 & Pick up the alphabet soup and place it in the basket. & $99.50\pm0.22$ & $\mathbf{100.00\pm0.00}$ \\
1 & Pick up the cream cheese and place it in the basket. & $\mathbf{100.00\pm0.00}$ & $\mathbf{100.00\pm0.00}$ \\
2 & Pick up the salad dressing and place it in the basket. & $99.50\pm0.22$ & $\mathbf{100.00\pm0.00}$ \\
3 & Pick up the BBQ sauce and place it in the basket. & $97.60\pm0.56$ & $\mathbf{98.00\pm0.00}$ \\
4 & Pick up the ketchup and place it in the basket. & $99.10\pm0.18$ & $\mathbf{100.00\pm0.00}$ \\
5 & Pick up the tomato sauce and place it in the basket. & $99.70\pm0.15$ & $\mathbf{100.00\pm0.00}$ \\
6 & Pick up the butter and place it in the basket. & $99.70\pm0.15$ & $\mathbf{100.00\pm0.00}$ \\
7 & Pick up the milk and place it in the basket. & $99.80\pm0.13$ & $\mathbf{100.00\pm0.00}$ \\
8 & Pick up the chocolate pudding and place it in the basket. & $\mathbf{100.00\pm0.00}$ & $\mathbf{100.00\pm0.00}$ \\
9 & Pick up the orange juice and place it in the basket. & $99.50\pm0.22$ & $\mathbf{100.00\pm0.00}$ \\
\addlinespace[2pt]
\multicolumn{2}{@{}l}{Suite mean} & $99.44\pm0.06$ & $\mathbf{99.80\pm0.00}$ \\
\bottomrule
\end{tabular}
\end{table}

\FloatBarrier
\begin{table}[!htbp]
\centering
\small
\setlength{\tabcolsep}{4pt}
\renewcommand{\arraystretch}{1.08}
\settowidth{\appendixtaskwidth}{ID$\mathbf{100.00\pm0.00}$$\mathbf{100.00\pm0.00}$}
\setlength{\appendixtaskwidth}{\dimexpr\linewidth-\appendixtaskwidth-6\tabcolsep\relax}
\caption{\textbf{Task-level success rates (\%) of $\pi_{0.5}$ on LIBERO-Goal and LIBERO-10.}
Task scores are mean $\pm$ standard error over 10 evaluation seeds for each trained model.
Bold marks the higher mean for each task.}
\label{tab:app-libero-goal-long-tasks}
\begin{tabular}{@{}r>{\raggedright\arraybackslash}p{\appendixtaskwidth}rr@{}}
\toprule
ID & Task & Flow & HT \\
\midrule
\multicolumn{4}{@{}l}{\textbf{LIBERO-Goal}} \\
0 & Open the middle drawer of the cabinet. & $96.10\pm0.46$ & $\mathbf{99.40\pm0.16}$ \\
1 & Put the bowl on the stove. & $98.20\pm0.53$ & $\mathbf{98.50\pm0.48}$ \\
2 & Put the wine bottle on top of the cabinet. & $89.90\pm0.60$ & $\mathbf{90.30\pm1.16}$ \\
3 & Open the top drawer and put the bowl inside. & $92.50\pm0.48$ & $\mathbf{95.40\pm0.85}$ \\
4 & Put the bowl on top of the cabinet. & $99.40\pm0.16$ & $\mathbf{99.50\pm0.27}$ \\
5 & Push the plate to the front of the stove. & $\mathbf{98.80\pm0.36}$ & $98.30\pm0.37$ \\
6 & Put the cream cheese in the bowl. & $97.70\pm0.37$ & $\mathbf{98.70\pm0.37}$ \\
7 & Turn on the stove. & $\mathbf{100.00\pm0.00}$ & $\mathbf{100.00\pm0.00}$ \\
8 & Put the bowl on the plate. & $98.90\pm0.31$ & $\mathbf{99.70\pm0.15}$ \\
9 & Put the wine bottle on the rack. & $\mathbf{94.00\pm0.94}$ & $89.60\pm0.86$ \\
\addlinespace[2pt]
\multicolumn{2}{@{}l}{Suite mean} & $96.55\pm0.19$ & $\mathbf{96.94\pm0.23}$ \\
\midrule
\multicolumn{4}{@{}l}{\textbf{LIBERO-10}} \\
0 & Living room 2: Put both the alphabet soup and the tomato sauce in the basket. & $\mathbf{94.50\pm0.54}$ & $92.00\pm0.00$ \\
1 & Living room 2: Put both the cream cheese box and the butter in the basket. & $\mathbf{98.30\pm0.26}$ & $96.00\pm0.00$ \\
2 & Kitchen 3: Turn on the stove and put the moka pot on it. & $96.80\pm0.39$ & $\mathbf{97.90\pm0.38}$ \\
3 & Kitchen 4: Put the black bowl in the bottom drawer of the cabinet and close it. & $99.00\pm0.33$ & $\mathbf{99.40\pm0.16}$ \\
4 & Living room 5: Put the white mug on the left plate and put the yellow and white mug on the right plate. & $97.60\pm0.16$ & $\mathbf{100.00\pm0.00}$ \\
5 & Study 1: Pick up the book and place it in the back compartment of the caddy. & $\mathbf{99.30\pm0.30}$ & $98.50\pm0.34$ \\
6 & Living room 6: Put the white mug on the plate and put the chocolate pudding to the right of the plate. & $\mathbf{94.80\pm0.59}$ & $94.00\pm0.00$ \\
7 & Living room 1: Put both the alphabet soup and the cream cheese box in the basket. & $96.40\pm0.34$ & $\mathbf{98.00\pm0.00}$ \\
8 & Kitchen 8: Put both moka pots on the stove. & $\mathbf{80.10\pm1.43}$ & $69.40\pm0.87$ \\
9 & Kitchen 6: Put the yellow and white mug in the microwave and close it. & $\mathbf{95.50\pm0.58}$ & $89.20\pm0.71$ \\
\addlinespace[2pt]
\multicolumn{2}{@{}l}{Suite mean} & $\mathbf{95.23\pm0.14}$ & $93.44\pm0.06$ \\
\bottomrule
\end{tabular}
\end{table}

\FloatBarrier

\subsubsection{Task-Level Results for GR00T N1.7}
\label{sec:appendix-gr1-results}

Tables~\ref{tab:appendix-gr1-results}, \ref{tab:appendix-bridge-results},
and~\ref{tab:appendix-fractal-results}
compare Flow-Policies and HT-Policies on RoboCasa-GR1, SIMPLER / Bridge,
and SIMPLER / Fractal, with officially reported Flow-Policy scores as reference.
These experiments fine-tune GR00T N1.7 from pretrained action-head
weights. The official Flow-Policy averages match those in
Table~\ref{tab:rq2-performance}.
We report Flow-Policy and HT-Policy scores as mean $\pm$ standard error across
10 evaluation seeds for each trained model, matching
Table~\ref{tab:rq2-performance}.

\begin{table}[!htbp]
\centering
\small
\setlength{\tabcolsep}{6pt}
\renewcommand{\arraystretch}{1.1}
\caption{\textbf{Task-level success rates (\%) on RoboCasa-GR1.}
Flow and HT report mean $\pm$ standard error across 10 evaluation
seeds for each trained model. Bold marks the higher mean of Flow and HT.}
\label{tab:appendix-gr1-results}
\begin{tabular}{@{}lrrr@{}}
\toprule
Task & Flow (official) & Flow & HT \\
\midrule
PnPBottle $\to$ CabinetClose & 70.00 & $42.00\pm4.23$ & $\mathbf{58.26\pm5.53}$ \\
PnPCan $\to$ DrawerClose & 70.00 & $23.90\pm2.68$ & $\mathbf{50.64\pm4.51}$ \\
PnPCup $\to$ DrawerClose & 35.00 & $8.40\pm1.83$ & $\mathbf{23.58\pm3.32}$ \\
PnPMilk $\to$ MicrowaveClose & 45.00 & $11.00\pm2.56$ & $\mathbf{25.00\pm4.22}$ \\
PnPPotato $\to$ MicrowaveClose & 40.00 & $6.00\pm1.25$ & $\mathbf{26.31\pm3.84}$ \\
PnPWine $\to$ CabinetClose & 65.00 & $24.88\pm2.98$ & $\mathbf{49.74\pm1.60}$ \\
\addlinespace[3pt]
Cuttingboard $\to$ Basket & 10.00 & $\mathbf{78.50\pm3.50}$ & $73.00\pm3.18$ \\
Cuttingboard $\to$ Cardboardbox & 30.00 & $31.32\pm2.10$ & $\mathbf{40.50\pm2.63}$ \\
Cuttingboard $\to$ Pan & 40.00 & $\mathbf{83.50\pm3.25}$ & $78.07\pm2.52$ \\
Cuttingboard $\to$ Pot & 45.00 & $\mathbf{73.50\pm2.69}$ & $72.50\pm3.27$ \\
Cuttingboard $\to$ Tieredbasket & 25.00 & $28.61\pm3.42$ & $\mathbf{34.50\pm1.89}$ \\
\addlinespace[3pt]
Placemat $\to$ Basket & 40.00 & $63.50\pm2.99$ & $\mathbf{67.50\pm2.71}$ \\
Placemat $\to$ Bowl & 40.00 & $35.40\pm2.48$ & $\mathbf{44.00\pm2.45}$ \\
Placemat $\to$ Plate & 40.00 & $\mathbf{42.50\pm3.52}$ & $41.00\pm2.33$ \\
Placemat $\to$ Tieredshelf & 18.20 & $12.06\pm2.47$ & $\mathbf{20.49\pm2.42}$ \\
\addlinespace[3pt]
Plate $\to$ Bowl & 50.00 & $39.00\pm2.33$ & $\mathbf{42.50\pm3.27}$ \\
Plate $\to$ Cardboardbox & 35.00 & $\mathbf{35.40\pm2.48}$ & $32.71\pm4.16$ \\
Plate $\to$ Pan & 40.00 & $40.71\pm3.40$ & $\mathbf{48.76\pm3.32}$ \\
Plate $\to$ Plate & 75.00 & $\mathbf{76.21\pm3.34}$ & $73.00\pm3.09$ \\
\addlinespace[3pt]
Tray $\to$ Cardboardbox & 60.00 & $67.05\pm4.03$ & $\mathbf{70.50\pm2.93}$ \\
Tray $\to$ Plate & 50.00 & $67.50\pm3.44$ & $\mathbf{72.00\pm3.89}$ \\
Tray $\to$ Pot & 45.00 & $69.50\pm4.25$ & $\mathbf{76.50\pm2.89}$ \\
Tray $\to$ Tieredbasket & 55.00 & $39.81\pm4.38$ & $\mathbf{53.00\pm2.60}$ \\
Tray $\to$ Tieredshelf & 45.00 & $15.81\pm3.66$ & $\mathbf{26.17\pm4.24}$ \\
\midrule
Mean & 44.50 & $42.34\pm0.68$ & $\mathbf{50.01\pm0.51}$ \\
\bottomrule
\end{tabular}
\end{table}

\begin{table}[!htbp]
\centering
\caption{\textbf{Task-level success rates (\%) on SIMPLER / Bridge.}
Flow and HT: mean $\pm$ standard error over 10 evaluation seeds.
Mean averages seven tasks; bold marks the higher mean of Flow and HT.}
\label{tab:appendix-bridge-results}
\small
\setlength{\tabcolsep}{6pt}
\renewcommand{\arraystretch}{1.1}
\begin{tabular*}{\linewidth}{@{\extracolsep{\fill}}lrrr@{}}
\toprule
Task & Flow (official) & Flow & HT \\
\midrule
carrot on plate & 58.00 & $\mathbf{63.00\pm1.94}$ & $29.80\pm1.17$ \\
close drawer & 97.00 & $98.40\pm0.50$ & $\mathbf{98.80\pm0.53}$ \\
put eggplant in basket & 53.00 & $50.80\pm2.33$ & $\mathbf{93.03\pm1.24}$ \\
put eggplant in sink & 2.00 & $2.00\pm0.84$ & $\mathbf{65.00\pm1.82}$ \\
open drawer & 100.00 & $\mathbf{100.00\pm0.00}$ & $72.60\pm1.19$ \\
spoon on towel & 78.00 & $\mathbf{87.02\pm1.13}$ & $61.40\pm1.98$ \\
stack cube & 48.00 & $\mathbf{33.20\pm1.50}$ & $15.57\pm0.66$ \\
\midrule
Mean & 62.29 & $62.06\pm0.40$ & $\mathbf{62.31\pm0.65}$ \\
\bottomrule
\end{tabular*}
\end{table}

\FloatBarrier
\begin{table}[!htbp]
\centering
\caption{\textbf{Task-level success rates (\%) on SIMPLER / Fractal.}
Flow and HT: mean $\pm$ standard error over 10 evaluation seeds.
Mean averages six tasks; bold marks the higher mean of Flow and HT.}
\label{tab:appendix-fractal-results}
\small
\setlength{\tabcolsep}{6pt}
\renewcommand{\arraystretch}{1.1}
\begin{tabular*}{\linewidth}{@{\extracolsep{\fill}}lrrr@{}}
\toprule
Task & Flow (official) & Flow & HT \\
\midrule
pick coke can & 100.00 & $\mathbf{97.90\pm0.41}$ & $97.80\pm0.57$ \\
pick object & 94.00 & $\mathbf{90.11\pm0.98}$ & $72.30\pm1.01$ \\
move near & 100.00 & $\mathbf{98.50\pm0.50}$ & $93.10\pm0.87$ \\
open drawer & 65.00 & $\mathbf{52.10\pm1.21}$ & $46.10\pm1.33$ \\
close drawer & 69.00 & $58.84\pm0.97$ & $\mathbf{66.20\pm1.19}$ \\
place in closed drawer & 7.00 & $9.50\pm1.01$ & $\mathbf{16.00\pm0.54}$ \\
\midrule
Mean & 72.50 & $\mathbf{67.83\pm0.39}$ & $65.25\pm0.44$ \\
\bottomrule
\end{tabular*}
\end{table}

\subsubsection{Results for Cosmos3-Nano on LIBERO-10}
\label{sec:appendix-cosmos-results}

Table~\ref{tab:appendix-cosmos-results} reports task-level success as
mean $\pm$ standard error across 10 evaluation seeds for each trained
model, using ten rollouts per task and seed. The overall results
match Table~\ref{tab:rq2-performance}.

\begin{table}[!htbp]
\centering
\caption{\textbf{Task-level success rates (\%) of Cosmos3-Nano on LIBERO-10.}
Mean $\pm$ standard error over 10 evaluation seeds; bold marks the higher mean per row.}
\label{tab:appendix-cosmos-results}
\small
\setlength{\tabcolsep}{4pt}
\renewcommand{\arraystretch}{1.1}
\begin{tabularx}{\linewidth}{@{}r>{\raggedright\arraybackslash}Xrr@{}}
\toprule
ID & Task & Flow & HT \\
\midrule
0 & Put both the alphabet soup and the tomato sauce in the basket. & $92.00\pm4.16$ & $\mathbf{95.00\pm2.24}$ \\
1 & Put both the cream cheese box and the butter in the basket. & $\mathbf{100.00\pm0.00}$ & $\mathbf{100.00\pm0.00}$ \\
2 & Turn on the stove and put the moka pot on it. & $94.00\pm2.21$ & $\mathbf{99.00\pm1.00}$ \\
3 & Put the black bowl in the bottom drawer of the cabinet and close it. & $94.00\pm2.21$ & $\mathbf{97.00\pm1.53}$ \\
4 & Put the white mug on the left plate and put the yellow and white mug on the right plate. & $95.00\pm1.67$ & $\mathbf{96.00\pm2.21}$ \\
5 & Pick up the book and place it in the back compartment of the caddy. & $\mathbf{99.00\pm1.00}$ & $98.00\pm1.33$ \\
6 & Put the white mug on the plate and put the chocolate pudding to the right of the plate. & $\mathbf{92.00\pm3.27}$ & $88.00\pm2.00$ \\
7 & Put both the alphabet soup and the cream cheese box in the basket. & $\mathbf{99.00\pm1.00}$ & $\mathbf{99.00\pm1.00}$ \\
8 & Put both moka pots on the stove. & $\mathbf{97.00\pm2.13}$ & $96.00\pm1.63$ \\
9 & Put the yellow and white mug in the microwave and close it. & $94.00\pm2.21$ & $\mathbf{97.00\pm1.53}$ \\
\midrule
\multicolumn{2}{@{}l}{Mean} & $95.60\pm0.90$ & $\mathbf{96.50\pm0.45}$ \\
\bottomrule
\end{tabularx}
\end{table}

\FloatBarrier

\subsection{Training Curves and Inference Timing}
\label{sec:appendix-efficiency}

\paragraph{Training-progress measurements.}
Figure~\ref{fig:convergence} plots success against optimizer
updates, rather than elapsed training time.
GR1 is evaluated after 2k, 16k, 32k, and 60k updates;
WidowX is evaluated after 5k, 10k, 15k, and 20k updates.
Each evaluation uses a single seed,
so these scores may differ from the 10-seed means in
Table~\ref{tab:rq2-performance}.
The Transport runs use state observations, absolute
actions, Chi-UNet, eight-step observation history, and eight
executed actions. They train for 300k updates and evaluate
every 20k updates, with 40 completed episodes per evaluation.
Transport-PH has three seeds per method. Transport-MH has
three Flow-Policy seeds, two MSE-Policy seeds, and seven HT-Policy seeds.
The Transport curves average scores across seeds at each training stage.
Flow-Policies use nine Euler steps for these Transport evaluations.

\paragraph{Hardware and measurement setup.}
Table~\ref{tab:inference-efficiency} reports batch-one inference on an
NVIDIA RTX 5090; Table~\ref{tab:app-inference-a800} reports additional
NVIDIA A800-SXM4-80GB measurements.
Measurements use the same architectures as the policy evaluations,
with BF16 mixed precision and no gradient computation or compilation.
Latency measures a policy query; throughput is
$1000/\mathrm{latency}_{\mathrm{ms}}$, measured in action chunks per second.

\begin{table}[htbp]
\centering
\caption{\textbf{Inference efficiency on an NVIDIA A800.}
Batch size one.
All latency values are reported in milliseconds.
NFE denotes the number of action-head evaluations per action chunk.
Speedup is the Flow-to-HT whole-model latency ratio.}
\label{tab:app-inference-a800}

\small
\setlength{\tabcolsep}{5.5pt}
\renewcommand{\arraystretch}{1.08}

\begin{tabular*}{\linewidth}{
@{\extracolsep{\fill}}
ll
c
c
c
c
c
@{}
}
\toprule
& &
\multicolumn{2}{c}{Action head} &
\multicolumn{3}{c}{Whole model} \\
\cmidrule(lr){3-4}
\cmidrule(lr){5-7}

Model
& Method
& NFE$\downarrow$
& Latency$\downarrow$
& Latency$\downarrow$
& FPS$\uparrow$
& Speedup$\uparrow$ \\
\midrule

\multirow{2}{*}{GR00T N1.7}
& Flow
& 4
& 89.2
& 139.1
& 7.19
& \multirow{2}{*}{\textbf{1.94$\times$}} \\

& HT (ours)
& \textbf{1}
& \textbf{24.3}
& \textbf{71.7}
& \textbf{13.95}
& \\

\addlinespace[2.5pt]

\multirow{2}{*}{$\pi_{0.5}$}
& Flow
& 10
& 169.7
& 260.8
& 3.83
& \multirow{2}{*}{\textbf{2.42$\times$}} \\

& HT (ours)
& \textbf{1}
& \textbf{17.1}
& \textbf{107.6}
& \textbf{9.29}
& \\

\addlinespace[2.5pt]

\multirow{2}{*}{Cosmos3-Nano}
& Flow
& 30
& 1368.2
& 1377.5
& 0.73
& \multirow{2}{*}{\textbf{16.74$\times$}} \\

& HT (ours)
& \textbf{1}
& \textbf{72.3}
& \textbf{82.3}
& \textbf{12.15}
& \\

\bottomrule
\end{tabular*}
\end{table}

\paragraph{Repeated measurements.}
For A800 measurements, methods within each model use the same 16 observations.
After 20 warmup calls, we time ten repetitions per observation
and report the mean over 160 synchronized measurements.

\Needspace{6\baselineskip}
\paragraph{Timing boundaries.}
Whole-model latency includes observation encoding and action prediction
from preprocessed GPU inputs. It excludes input/output processing,
environment execution, and video decoding. Head latency measures
action prediction with observation features already computed.
GR00T and $\pi_{0.5}$ skip scale prediction when evaluating HT-Policies.

\paragraph{Cosmos3-Nano timing configuration.}
For Cosmos3-Nano, the Flow-Policy uses 30 sampling steps, while the
HT-Policy predicts actions in a single forward pass through the joint
video/action network.

\Needspace{16\baselineskip}
\subsection{Additional Ablations and Settings}
\label{sec:appendix-loss-settings}

\subsubsection{Regression Objectives}
\label{sec:residual-model-ablation}

\begin{wraptable}{r}{0.42\textwidth}
    \centering
    \captionsetup{font=small,skip=4pt,justification=raggedright,singlelinecheck=false}
    \caption{
    \textbf{Regression loss ablation.}
    GR00T N1.7 on RoboCasa-GR1, 60k steps, 24 tasks.
    Upper/lower blocks omit/use input-dependent scales.
    Both Student-$t$ variants use $\nu=928$.
    }
    \label{tab:robust-loss-ablation}
\begingroup
      \centering
      \small
      \renewcommand{\arraystretch}{1.08}
      \begin{tabularx}{\linewidth}{@{}>{\raggedright\arraybackslash}Xr@{}}
          \toprule
          Objective & SR (\%) $\uparrow$ \\
          \midrule
          MSE                           & 37.80 \\
          $L_1$                         & 27.00 \\
          RMSE                          & 46.00 \\
          Huber~\cite{Huber_1964}                         & 38.10 \\
          Student-$t$                   & 39.50 \\
          \midrule
          Heteroscedastic Gaussian      & 42.50 \\
          Heteroscedastic $L_1$         & 44.50 \\
          Heteroscedastic RMSE           & 47.67 \\
          Heteroscedastic Huber         & 40.70 \\
          Heteroscedastic Student-$t$ (HT) & \textbf{51.50} \\
          \bottomrule
      \end{tabularx}\par
\endgroup

\end{wraptable}

\paragraph{Scale adaptation and robust losses.}
The objective ablations in Table~\ref{tab:robust-loss-ablation} show
that learning an observation-dependent scale improves success for
every tested objective.
We attribute this consistent benefit to the heteroscedastic residuals
observed in Section~\ref{sec:tail-origin}, whose typical magnitudes
vary across observations.
Different robust objectives~\citep{barron2019generaladaptiverobustloss} also improve over MSE, with heteroscedastic
Student-$t$ achieving the highest success.
The only exception is fixed-scale $L_1$.
$L_1$ penalizes coordinates independently, while RMSE uses a joint
norm that makes each coordinate's gradient depend on the full
residual vector.
Replacing $L_1$ with RMSE improves success under both fixed and
learned scales, supporting the value of treating the action residual
jointly.

\paragraph{Common training setting.}
The objective comparison uses GR00T N1.7 on RoboCasa-GR1 after
60,000 updates, with the data, mask, and optimizer described
in Appendix~\ref{sec:appendix-training}.
Success rates in Table~\ref{tab:robust-loss-ablation} use a single
evaluation seed per model, so they may differ from the 10-seed
means in Table~\ref{tab:rq2-performance}.
Every action chunk has $d=232$ valid coordinates.
Both Student-$t$ variants use $\nu=928$.
Table~\ref{tab:app-loss-definitions} specifies the implemented
penalties. Scale-dependent terms are included during training;
only constants independent of the optimized parameters are omitted.

\begin{table}[htbp]
\centering
\caption{\textbf{Implemented regression penalties.} Penalties are shown before
division by the number of valid action coordinates. Let $S=\|r\|_2^2$,
$R=\|r\|_2$, $Q=\|r\|_1$, and
$T_\nu(z)=\frac{\nu+d}{2}\log(1+z^2/\nu)$.
The learned scale $\sigma=\sigma(o)$ is scalar.}
\label{tab:app-loss-definitions}
\small
\setlength{\tabcolsep}{6pt}
\renewcommand{\arraystretch}{1.4}
\begin{tabularx}{\linewidth}{@{}lXl@{}}
\toprule
Family & Fixed-scale penalty & Learned-scale penalty \\
\midrule
Gaussian & $S$ & $S/(2\sigma^2)+d\log\sigma$ \\
$L_1$ & $Q$ & $Q/\sigma+d\log\sigma$ \\
Radial norm (RMSE) & $R$ & $R/\sigma+d\log\sigma$ \\
Huber & $h_\delta(R)$ & $h_\delta(R/\sigma)+d\log\sigma$ \\
Student-$t$ & $T_\nu(R/\sigma_0)+d\log\sigma_0$ & $T_\nu(R/\sigma)+d\log\sigma$ \\
\bottomrule
\end{tabularx}
\end{table}

\paragraph{Coordinate and radial penalties.}
$L_1$ sums absolute errors independently across coordinates.
The norm and Huber losses act on the whole chunk radius.
The ``RMSE'' label denotes the radial-norm baseline: its
implementation minimizes $R/d$, which differs from the
conventional $R/\sqrt d$ by a fixed factor for this experiment.
The corresponding likelihood is a spherical radial-exponential
density, rather than a product of coordinate-wise Laplace densities.
Huber uses $h_\delta(z)=z^2/2$ for $z\leq\delta$ and
$h_\delta(z)=\delta z-\delta^2/2$ otherwise.
Its threshold is 2.5 times an exponential moving average of
the batch median radius, with decay 0.99. The heteroscedastic
version applies the same rule to the standardized radius.

\paragraph{Scale initialization.}
Fixed-scale Student-$t$ uses $\sigma_0\simeq0.291346$.
Heteroscedastic RMSE (Hetero-RMSE) requires a different scale convention from Gaussian
or Student-$t$ regression. We multiply its softplus scale, including
the $10^{-3}$ floor, by a constant chosen on the initial training
batch so that its mean equals $\E[R]/d$.
The scale head remains trainable, while this multiplier stays fixed.
This initialization is separate from the post-training
likelihood calibration in Appendix~\ref{sec:likelihood-calibration}.

\FloatBarrier
\subsubsection{Likelihood Calibration}
\label{sec:likelihood-calibration}

The success rates paired with likelihoods in
Figure~\ref{fig:dist-quality-vs-sr} use a single evaluation seed per model,
whereas Table~\ref{tab:rq2-performance} averages 10 evaluation seeds.

\paragraph{A single global scale multiplier.}
For each frozen policy, we replace its residual scale $\sigma(o)$
with $g\sigma(o)$ and choose
\begin{equation}
    g^*=\arg\max_{g>0}\E_{\mathrm{cal}}
    \left[\log p\big(a\mid\mu(o),g\sigma(o)\big)\right].
    \label{eq:global-scale-calibration}
\end{equation}
We use the full joint log-likelihood with all log-scale terms and
normalization constants, using radial densities for RMSE and Huber.
One $g$ is shared across all tasks, observations, and coordinates,
preserving relative scales. Policy weights, distribution family, $\nu$,
and Huber thresholds remain fixed. Calibration neither uses nor changes
policy success rates.

\paragraph{Calibration and scoring data.}
We sample 512 demonstration episodes per GR1 task, totaling 12,288.
Each contributes one eight-step chunk with 232 valid coordinates and
four forward passes with training augmentation and dropout.
We use four task-stratified folds, grouping all passes from each episode
together. Each fold is scored with $g$ fitted on the other three folds.
These demonstrations were used in policy training; the held-out split
applies only to scale calibration.

\paragraph{Reported likelihood.}
The horizontal axis in Figure~\ref{fig:dist-quality-vs-sr} is
$\E[\log p(a\mid o)]/232$, the mean joint log-likelihood per action
dimension in nat/dim, averaged over forward passes and held-out folds.
Passes are scored separately and do not define a mixture density.
The Gaussian special case has
$g^2=\E_{\mathrm{cal}}[\|r\|_2^2/\sigma(o)^2]/d$.
For fixed-scale MSE, the calibrated standard deviation is
the calibration residual RMS.

\paragraph{Huber thresholds and uncertainty.}
For likelihood calibration, the Huber threshold is fixed at
1.87205 for fixed-scale Huber and 34.1826 for heteroscedastic
Huber (Hetero-Huber).
Confidence intervals use 2,000 task-stratified episode bootstrap samples
with each fitted $g$ fixed. They reflect episode sampling, rather than
variation across training seeds or control rollouts.

\subsubsection{Degrees-of-Freedom Sweep Settings}
\label{sec:appendix-nu-settings}

Table~\ref{tab:app-nu-grid} lists the $\nu$ values and training
budgets for the points shown in Figure~\ref{fig:nu-sensitivity}.
Each point uses a single evaluation seed, so its success rate may differ
from the 10-seed mean in Table~\ref{tab:rq2-performance}.
All variants learn observation-dependent scales.
The GR1 and Bridge runs follow the corresponding GR00T recipes;
the $\pi_{0.5}$ runs use 30k updates on all four LIBERO suites.
All Bridge evaluations use the environment's default random
initialization.

\begin{table}[htbp]
\centering
\caption{\textbf{Degrees-of-freedom settings for Figure~\ref{fig:nu-sensitivity}.}
Bold marks the
choice $\nu=4d$ ($c=4$), which gives the highest tested success
in each setting. HG denotes the Gaussian limit.}
\label{tab:app-nu-grid}
\small
\renewcommand{\arraystretch}{1.1}
\begin{tabularx}{\linewidth}{@{}lrrX@{}}
\toprule
Setting & Updates & $d$ & Reported $\nu$ \\
\midrule
RoboCasa-GR1 & 60,000 & 232 & $2,116,464,\mathbf{928},1450,\infty$ (HG) \\
Bridge & 20,000 & 56 & $56,112,\mathbf{224},350,448$ \\
$\pi_{0.5}$ / LIBERO & 30,000 & 350 & $700,\mathbf{1400},2187.5$ \\
\bottomrule
\end{tabularx}
\end{table}

\FloatBarrier

\FloatBarrier
\endgroup

\end{document}